%% file: main.tex
\documentclass[11pt]{article}

\usepackage[preprint]{acl}

\usepackage{times}
\usepackage{latexsym}
\usepackage[T1]{fontenc}
\usepackage[utf8]{inputenc}
\usepackage{microtype}
\usepackage{inconsolata}

\usepackage{graphicx}
\usepackage{hyperref}
\usepackage{arydshln}
\usepackage{algorithm}
\usepackage{algpseudocode}
\usepackage[skins,breakable,most]{tcolorbox}
\usepackage{enumitem}
\usepackage{makecell}
\usepackage{booktabs}
\usepackage{multirow}
\usepackage{amsmath}
\usepackage{amssymb}
\usepackage{cleveref}
\usepackage{subcaption}
\usepackage{array}
\usepackage{adjustbox}
\usepackage{soul}
\usepackage{caption}
\usepackage{svg}
\usepackage{placeins}
\usepackage{xurl}
\usepackage{tabularx}
\usepackage{xcolor}
\usepackage{colortbl}

\usepackage{annotation}

\usepackage{tikz}
\usetikzlibrary{arrows.meta,positioning,calc}

\input{prompt_box_style}
\input{macros}

\crefname{tcb@cnt@promptbox}{Prompt}{Prompts}
\Crefname{tcb@cnt@promptbox}{Prompt}{Prompts}

\definecolor{pastelBlue}{RGB}{174,198,207}
\definecolor{pastelRed}{RGB}{255,179,186}
\definecolor{pastelOrange}{RGB}{255,223,186}
\definecolor{pastelGreen}{RGB}{203,240,191}
\definecolor{pastelViolet}{RGB}{213,181,226}
\definecolor{pastelYellow}{RGB}{255,255,204}
\definecolor{pastelPurple}{RGB}{230,230,250}
\definecolor{pastelPink}{RGB}{255,192,203}

\title{Speak for Me: Giving LLMs the Situational Awareness\\to Participate in a Meeting}

\author{
  Muneeb Khan%
  \thanks{Equal contribution;
  author order is alphabetical.},
  Frederic Kirstein%
  \footnotemark[1]%
  \thanks{Corresponding author.},
  Terry Ruas, Bela Gipp \\
  University of Göttingen, Germany \\
  \texttt{frederic.kirstein@uni-goettingen.de}
}

\begin{document}
\maketitle
\AddAnnotationRef

\begin{abstract}
\input{text_short/abstract}
\end{abstract}

\section{Introduction} \label{sec:introduction}

\input{text_short/01_introduction}

\section{Related Work} \label{sec:related_work}
\input{text_short/02_related_work}

\section{Methodology} \label{sec:methodology}
\input{text_short/03_methodology}

\section{Experimental Setup} \label{sec:experimental_setup}
\input{text_short/04_experimental_setup}

\section{Results} \label{sec:results}
\input{text_short/05_results}

% \clearpage

\section{Ablations} \label{sec:ablations}
\input{text_short/06_ablations}

\FloatBarrier

\section{Final Considerations} \label{sec:conclusion}
\input{text_short/06_conclusion}

\section*{Limitations} \label{sec:limitations}
\input{text_short/07_limitations}

\section*{Acknowledgements}
This work was supported by the Lower Saxony Ministry of Science and Culture and the VW Foundation.

\bibliography{custom}

\appendix

\section*{Ethical Considerations} \label{sec:ethics}
\input{text_short/08_ethics}

% --- Two-column appendix: resources, setup details, human eval ---
\input{text_short/B1_resources}
\input{text_short/B2_setup}
\input{text_short/A2_compute}
\input{text_short/B3_human_eval}
\input{text_short/A3_robustness}

% --- Single-column for system prompts (need full page width) ---
\input{text_short/A1_prompts}

\clearpage
\onecolumn
\input{annotation.tex}

\end{document}

%% file: prompt_box_style.tex
\usepackage{tcolorbox}
\tcbuselibrary{skins, breakable, listings}

\definecolor{promptTeal}{HTML}{006666}      % Header background
\definecolor{promptGray}{HTML}{F5F5F5}      % Content background
\definecolor{promptTextDark}{HTML}{333333}  % Body text

\newtcolorbox[auto counter, number within=section]{promptbox}[2][]{%
  enhanced,
  breakable,
  colback=promptGray,
  colframe=promptTeal,
  colbacktitle=promptTeal,
  coltitle=white,
  fonttitle=\bfseries\sffamily\small,
  title={\thetcbcounter: #2},
  sharp corners,
  boxrule=0.8pt,
  toptitle=1.5mm,
  bottomtitle=1.5mm,
  left=3mm,
  right=3mm,
  top=1.5mm,
  bottom=1.5mm,
  #1
}

\newtcblisting{promptcode}{%
  enhanced,
  breakable,
  colback=white,
  colframe=promptGray,
  boxrule=0.5pt,
  left=2mm,
  right=2mm,
  top=1mm,
  bottom=1mm,
  sharp corners,
  listing only,
  listing options={%
    basicstyle=\ttfamily\scriptsize,
    breaklines=true,
    breakatwhitespace=false,
    columns=flexible,
    keepspaces=true,
    showstringspaces=false,
    tabsize=2,
    xleftmargin=0pt,
  }
}

\newenvironment{prompttext}{%
  \par\vspace{2pt}%
  \small%
  \ttfamily%
  \setlength{\parindent}{0pt}%
  \setlength{\parskip}{3pt}%
}{%
  \par\vspace{2pt}%
}

\newtcolorbox{subpromptbox}[2][]{%
  enhanced,
  breakable,
  colback=promptGray!50,
  colframe=promptTeal!70,
  colbacktitle=promptTeal!70,
  coltitle=white,
  fonttitle=\bfseries\sffamily\small,
  title={#2},
  sharp corners,
  boxrule=0.6pt,
  toptitle=1.5mm,
  bottomtitle=1.5mm,
  left=3mm,
  right=3mm,
  top=1.5mm,
  bottom=1.5mm,
  grow to left by=3mm,
  grow to right by=3mm,
  #1
}

%% file: macros.tex
\usepackage{booktabs}
\usepackage{adjustbox}   % for \begin{adjustbox}{max width=...}
\usepackage{placeins}    % for \FloatBarrier (only if you use it)

\newcommand{\sysname}{CAPA}
\newcommand{\sysfull}{Collaborative Agent Predictive Architecture}
\providecommand{\estci}[2]{%
  \shortstack[c]{%
    \strut #1\strut\\[-0.15ex]
    {\scriptsize\color{black!80}[#2]}%
  }%
}

\definecolor{ourrow}{HTML}{EAF2FB}

%% file: text_short/abstract.tex
In online meeting delegation, LLM agents fail to recognize when to
speak. With no structured way to track stances, coverage, and floor,
they miss the moments where they should contribute. Prompt-only
delegates stay silent on $51.4\%$ of the absent participant's talking
opportunities on the AMI corpus.
We present \sysname{} (\sysfull{}), an architecture for online meeting
delegation. A Perceiver updates the meeting state from each observed
turn. A Predictor forecasts how the conversation will continue. A
Controller decides whether to speak and which proposition to surface.
A Generator phrases the chosen contribution in the participant's
style. Two judges score the forecast and the action against the next
observed turn. A Recalibrator updates the meeting state from those
verdicts for future decisions.
To evaluate online delegation, we introduce an episode-level protocol
that scores whether, when, and what a delegate contributes around the
participant's actual idea units. The protocol's schema-constrained
LLM judges align with human annotations at Cohen's
$\kappa = 0.71$.
On 137 AMI meetings, \sysname{} reduces the silence rate from $51.4\%$
to $2.5\%$, doubles credited recovery ($26.1\!\to\!52.2$), and keeps
hallucination at $0.6\%$. The failure mode shifts from omission to
selection, with each residual near-miss attributable to a specific
module of the architecture. Mechanism ablations identify the meeting
state as the lever that closes the recognition gap, where raw-context
scaling alone does not.\footnote{Supplementary material is available as according to \Cref{app:repository}.}

%% file: text_short/01_introduction.tex
As LLMs evolve into proactive agents, they must increasingly navigate multi-party environments where deciding \emph{whether} to intervene is as consequential as deciding what to say.
We examine this capability through participant-specific \emph{meeting delegation}, where an LLM proxy actively represents an absent stakeholder \citep{leong2024dittos,hu-etal-2025-meeting}.
In modern organizational settings, consistent attendance is increasingly difficult to guarantee \citep{allen2023_key_features_meetings, mok2023_timezones, microsoft2025_infiniteworkday}.
Post-hoc summarization and recordings offer retrospective review but do not allow an absent participant to \emph{intervene} as outcomes are actively negotiated \citep{nathan2012_incase, asthana2025_meetingrecap}.
Because decisions made without the relevant stakeholders are costly to reverse \citep{sleesman2012bigmuddy}, a proxy must be able to step into the live conversation at the right time.

\input{figures/02_system}

A delegate must continuously decide \emph{whether} to speak at each turn and, if so, \emph{which} contribution to make in the participant's voice.
Current prompt-based approaches fail at both \citep{hu-etal-2025-meeting}.
In a controlled replication on the AMI corpus \citep{carletta2005ami}, baselines remain silent on $51.4\%$ of the absent participant's valid talk opportunities, and two failure modes recur when they do interject.
The cue policy is bound to surface invitation, triggering on explicit name-call but missing implicit handoffs and role-specific opportunities for the represented participant.
The content policy conflates topical relevance with proposition-level coverage, i.e., suppressing valid contributions because the topic has been touched, and binds live references to the represented participant unreliably.

Both modes originate from an upstream limitation in state tracking.
Standard proactive agents do not explicitly model participant stances, unresolved issues, or floor control, and therefore fail to identify intervention windows or to ground responses in the immediate context.
Extending the input transcript does not resolve this, since longer context windows do not inherently resolve reasoning deficits over long conversations \citep{liu-etal-2024-lost, du-etal-2025-context}.
Delegation therefore relies on a structured representation of the meeting state, coupled with a mechanism to continuously synchronize it with the live conversation.

We introduce \textbf{\sysname{}} (\textit{\sysfull{}}), a fixed-weight architecture for multi-party delegation built on a \emph{perceive--act--recalibrate} loop (\Cref{fig:capa-architecture}).
\sysname{} maintains a structured \emph{meeting state} $\mathbf{m}_t$ that tracks six action-relevant fields derived from documented prompt-only failure modes \citep{hu-etal-2025-meeting}: active topic, decisions, open questions, participant stances, prior coverage, and floor.
We study this problem under a causal replay protocol. 
At each turn, \sysname{} observes only the dialogue prefix available at that time, updates $\mathbf{m}_t$, decides whether and how to intervene, and forecasts the next turn.
Two LLM judges then score the forecast and the delegate's action against the observed continuation, and the resulting verdicts feed back into $\mathbf{m}_t$, shifting self-correction from output refinement to continuous state maintenance.

Because standard generation metrics score turn-by-turn text overlap and further do not consider timing, we introduce an \emph{episode-level} evaluation protocol anchored on the substantive contributions the absent stakeholder actually made, i.e., the participant-owned \emph{idea units} to obtain a controlled set of substantive intervention opportunities.
For each unit, programmatic checks score \emph{whether} and \emph{when} the delegate intervened, while schema-constrained LLM judges score \emph{what} it contributed and are validated against human annotations (\Cref{app:human-eval}).
Evaluated under this protocol across 137 AMI meetings, \sysname{} reduces the silence rate from $51.4\%$ to $2.5\%$, doubles credited recovery ($26.1\!\to\!52.2$), and keeps hallucination at $0.6\%$.
We conclude that (i) overcoming the dominant ``silent abstention'' failure in multi-party agents requires an explicit, structured meeting state, as raw-context scaling alone cannot resolve long-range opportunity recognition; (ii) the bottleneck for LLM delegates lies in the binary floor-taking decision rather than generation timing, an issue that explicit state tracking successfully mitigates; and (iii) transitioning from monolithic prompt-based generation to a modular, state-driven architecture transforms opaque omission failures into transparent, module-attributable selection errors.

\paragraph{Contributions.}
\begin{itemize}[leftmargin=*,nosep]
\item \sysname{}, a perceive--act--recalibrate based causal reply protocol that shifts multi-party LLM delegation from context-reading at decision time to continuous state maintenance.
\item A unified evaluation protocol that scores delegate interventions against actual participant idea units, evaluating both timing and content alignment.
\item Empirical insights on the AMI corpus, where explicit state tracking shifts the failure mode from omission to bounded near-misses with localizable architectural causes, a result raw-context scaling does not reproduce.
\end{itemize}

%% file: figures/02_system.tex
\begin{figure*}[t]
\centering
\includegraphics[width=\linewidth]{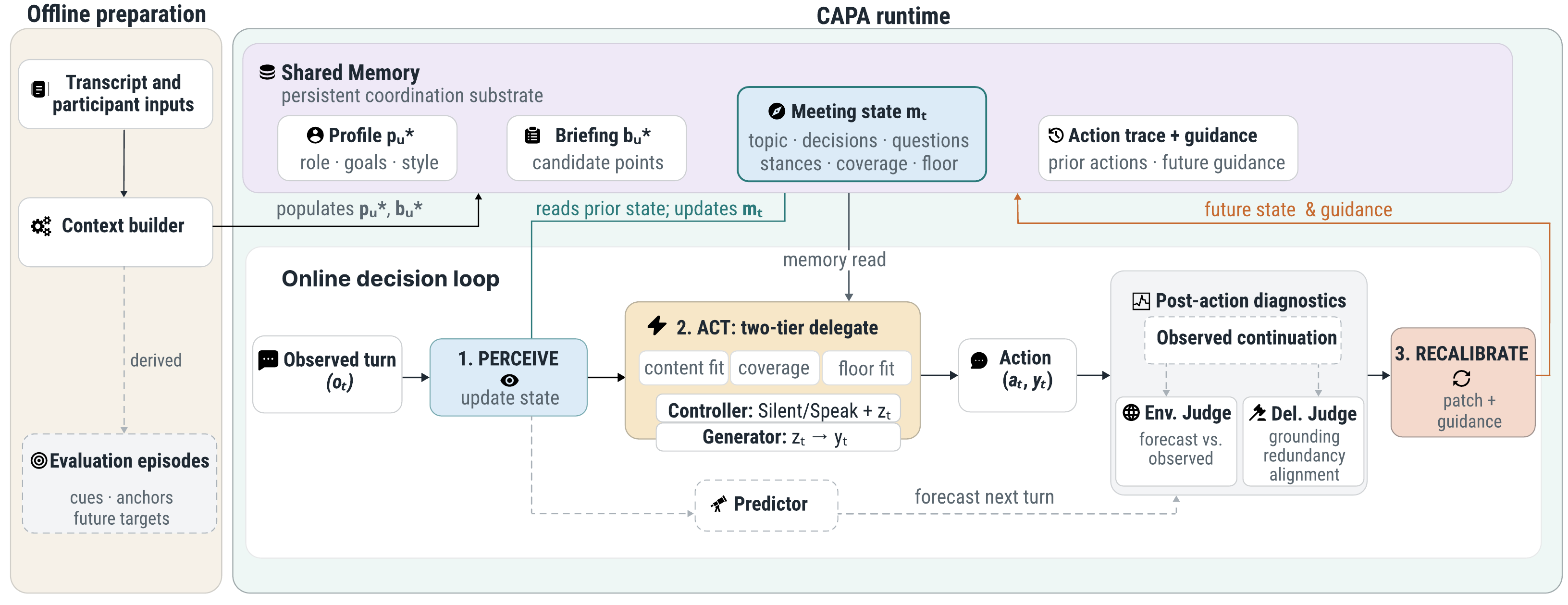}
% \caption{\textbf{\sysname{} architecture.}
% Offline preparation builds participant context and stores it in Shared
% Memory; fixed-window episodes are derived separately for evaluation, with
% future anchors and targets withheld from the runtime delegate. At each
% online decision, \textsc{Perceive} updates the explicit meeting state
% $\mathbf{m}_t$, \textsc{Act} reads memory to choose
% \textsc{Silent}/\textsc{Speak} and commit to a semantic point $z_t$
% before realizing $y_t$, and \textsc{Recalibrate} uses the Predictor and
% two independent judges to patch only future state and guidance.}
\caption{\sysname{} architecture.
Offline preparation builds participant context in Shared Memory; fixed-window
episodes are derived separately for evaluation, with future anchors and
targets withheld from the delegate. At each online decision,
\textsc{Perceive} updates $\mathbf{m}_t$, \textsc{Act} reads memory to
choose \textsc{Silent}/\textsc{Speak} and commit to $z_t$ before realizing
$y_t$, and \textsc{Recalibrate} uses the Predictor and two independent
judges to patch future state and guidance.}
\label{fig:capa-architecture}
\end{figure*}

%% file: text_short/02_related_work.tex
\paragraph{Meeting agents.}
Current LLM applications in meeting domains primarily serve as passive observers or post hoc facilitators, focusing on summarization, leading, and question answering \citep{mao2024muca, alsobay2025gpt4moderator, asthana2025_meetingrecap, chen2025_meetmap, jacniacki2025_huma, sapkota2025_mpcas_survey}.
While these systems support retrospective review, they do not participate continuously or represent specific stakeholders during live interactions.
Although user studies indicate demand for real-time proxies \citep{leong2024dittos, nathan2012_incase}, building such systems remains a nascent computational challenge.
\citet{hu-etal-2025-meeting} introduce the closest benchmark to our setting, documenting failure modes that arise when delegation relies exclusively on prompting.
In contrast to work focused solely on elicitation, \sysname{} introduces the architectural substrate required for online intervention, with explicit state management as the load-bearing component.

\paragraph{Sequential decision-making under partial observability.}
Delegation has the defining structure of sequential decision-making under partial observability. 
Decisions are coupled because each \textsc{Speak} or \textsc{Silent} choice reshapes the conversation and the remaining opportunities.
The decision-relevant state is latent because stances, coverage, and floor must be inferred.
The objective is episodic, scoring whether, when, and what the delegate contributes.
The classical response to partial observability is to act on a belief over the latent state \citep{kaelbling1998pomdp}.
POMDP-based dialogue management applies this decomposition to task-oriented slot filling with learned policies \citep{williams2007pomdp,young2013pomdp}, rather than proactive multi-party intervention \citep{deng2023proactive}.
\sysname{} adopts the belief-and-policy decomposition and instantiates it with schema-constrained LLM modules.

\paragraph{Inference-time correction.}
Post-action self-correction has emerged as a prevailing inference-time strategy for steering fixed-weight agents.
Paradigms such as Reflexion \cite{shinn2023_reflexion}, Self-Refine \cite{ madaan2023_selfrefine}, and their variants rely on critiquing a generated utterance and iteratively rewriting it for the next attempt \cite{pan-etal-2024-correcting, ma-etal-2025-improving, phan-etal-2025-think}.
Similarly, recent work leverages interaction dynamics as a learning signal to refine subsequent outputs \citep{gooding2025_trace, hilgert2025_nextspeaker}.
Across these approaches, the correction channel strictly targets the \emph{linguistic output}.
\sysname{} instead directs feedback into the explicit \emph{meeting state} that conditions subsequent decisions.
Its recalibration step therefore also addresses silent failures for which no utterance exists to rewrite.

%% file: text_short/03_methodology.tex
\label{sec:methodology}

\subsection{Problem Formulation}
\label{subsec:problem-formulation}

We treat online meeting delegation as a sequential decision-making problem under partial observability.
The delegate represents an absent participant $u^\star$ and acts in their voice during a live meeting.
It conditions on a participant context $\mathbf{c}_{u^\star} = (\mathbf{p}_{u^\star}, \mathbf{b}_{u^\star})$, comprising a stable long-term profile $\mathbf{p}_{u^\star}$ (role, expertise, communication style) and a meeting-specific briefing $\mathbf{b}_{u^\star}$ (candidate contributions for the current meeting).
At each decision point, the delegate observes the dialogue history $H_t$ (the sequence of speaker--utterance pairs up to turn $t$) and chooses an action $a_t \in \{\textsc{Silent}, \textsc{Speak}\}$.
On \textsc{Speak}, it first commits to a discrete semantic proposition $z_t$ (a specific claim or question) and only then realizes it as a natural-language utterance.
The information that drives timing and grounding decisions, such as the active topic, unresolved decisions, open questions, participant stances, and floor, is not directly observable from $H_t$.
Approaches that condition generation directly on the raw dialogue history therefore exhibit contextual drift.

\subsection{Desiderata}
\label{subsec:desiderata}

\begin{table*}[t!]
\centering
\scriptsize
\setlength{\tabcolsep}{8pt}
\renewcommand{\arraystretch}{1.3}
\definecolor{capaHeader}{HTML}{1F2A44}
\definecolor{capaTaskTint}{HTML}{EAF2FB}
\definecolor{capaArchTint}{HTML}{EFE9F7}
\begin{tabularx}{\textwidth}{@{}>{\raggedright\arraybackslash}p{0.22\textwidth}X@{}}
\toprule
\rowcolor{capaHeader} \multicolumn{2}{l}{\textcolor{white}{\textsf{\textbf{Task-design constraints} }}} \\
\addlinespace[2pt]
\textbf{(i)\ Strict Causality} & Decisions at turn $t$ condition only on dialogue observed up to $t$. Dropping this invalidates the online setting. \\
\textbf{(ii)\ Participant Grounding} & Interventions are bounded by the expertise, goals, and briefing of $u^\star$. Without this, the proxy regresses to a generic LLM. \\
\textbf{(iii)\ Pre-generation Commitment} & The delegate selects a discrete proposition $z_t$ before surface realization, preventing hallucinations that arise when generation dictates reasoning. \\
\textbf{(iv)\ Abstention Sensitivity} & \textsc{Silent} is a first-class action. Missed opportunities are penalized symmetrically with inappropriate interruptions. \\
\midrule
\rowcolor{capaHeader} \multicolumn{2}{l}{\textcolor{white}{\textsf{\textbf{Architectural commitments} }}} \\
\addlinespace[2pt]
\textbf{(v)\ Explicit State Maintenance} & Action-relevant variables persist as an explicit meeting state $\mathbf{m}_t$. Without this, the agent loses situational context. \\
\textbf{(vi)\ State-Directed Correction} & Feedback from subsequent dialogue updates $\mathbf{m}_t$ for future decisions. Otherwise, misinterpretations accumulate as contextual drift. \\
\bottomrule
\end{tabularx}
\caption{Six requirements derived from prompt-only failure modes \citep{hu-etal-2025-meeting}, grouped into task-design constraints (i--iv) and architectural commitments (v--vi).}
\label{tab:desiderata}
\end{table*}

To overcome the representation deficit, a proxy architecture must satisfy a specific set of structural constraints.
We derive these desiderata directly from the empirical failure modes of zero-shot delegates, i.e., silent abstention, redundancy, off-topic hallucination, and mistiming \citep{hu-etal-2025-meeting}.
\Cref{tab:desiderata} states the six constraints we propose as a minimal and complete set relative to these known failures.
The first four govern the task design. The latter two define the required architectural paradigm.

\subsection{The \sysname{} Framework}
\label{subsec:capa}

To satisfy the desiderata (\Cref{tab:desiderata}), we introduce \sysname{} (\textit{\sysfull{}}), a continuous \emph{perceive--act--recalibrate} loop over an explicit meeting state $\mathbf{m}_t$ (see \Cref{fig:capa-architecture}).
The architecture decomposes the delegation task into four parts: a \texttt{Shared Memory} that holds the participant context and the evolving state, a \texttt{Perceiver} that estimates the state from each new turn, a two-tier \texttt{Controller--Generator} that selects and realizes an action, and a \texttt{Recalibrator} that integrates dual-judge feedback into the state.
Conceptually, $\mathbf{m}_t$ is an explicit, approximate belief over the decision-relevant interaction variables, updated by the \texttt{Perceiver} and acted on by the \texttt{Controller} \citep{kaelbling1998pomdp,williams2007pomdp}.
This POMDP-inspired belief--policy decomposition gives the modules a persistent, inspectable decision substrate and allows the \texttt{Recalibrator} to correct specific state fields after observed errors.
\sysname{} uses the decomposition as an architectural principle. Learning rewards, transition dynamics, or a planning policy would require a reliable meeting simulator and participant-specific interaction trajectories, which are unavailable in this setting.
By centralizing memory and separating perception, action, and correction, \sysname{} avoids the contextual drift that direct sequence-to-sequence mapping over raw dialogue introduces.
The rest of this section instantiates each part. 
Prompts and schemas are in \Cref{sec:appendix-prompts}.

\paragraph{Centralized Shared Memory.}
The \texttt{Shared Memory} centralizes all information the modules read or write.
We organize it along two orthogonal axes that capture the information lifecycle in a meeting: \emph{timescale} (stable across the meeting vs.\ changing turn by turn) and \emph{source} (brought in by the delegate vs.\ generated by the environment).
The resulting 2$\times$2 partition yields four schema-constrained stores. Long-term memory holds the stable internal profile $\mathbf{p}_{u^\star}$ (who the delegate is). Session memory holds the stable external briefing $\mathbf{b}_{u^\star}$ (what the delegate is here to do). Working memory holds the dynamic external state $\mathbf{m}_t$ (what is happening). Episodic memory holds the dynamic internal trace of prior actions.
This specific partition avoids the redundancy of finer splits, which would share identical update cycles, and the inefficiency of coarser aggregations, which would compel runtime filtering across mixed lifecycles.
Centralizing the four stores satisfies Desideratum ii because participant attributes persist across the whole meeting, and Desideratum v because action-relevant variables become fields the action and recalibration modules read directly.

\paragraph{Perceive.}
A single \texttt{Perceiver} step (\Cref{prompt:eim}) concentrates state estimation, preventing inference cost from scaling with the number of downstream consumers and avoiding cross-module inconsistency.
The \texttt{Perceiver} is a schema-constrained LLM call that maps the newly observed turn $\mathbf{o}_t$ and the prior state $\mathbf{m}_{t-1}$ to an updated $\mathbf{m}_t$.
This extends dialogue-state tracking from task slots to interaction variables such as stance, coverage, and floor \citep{young2013pomdp}.
The state has five fields, each addressing a prompt-only failure mode documented by \citet{hu-etal-2025-meeting}: active topic (off-topic), decisions and open questions (missed opportunities), prior coverage (redundancy), participant stances (participant grounding), and floor (mistiming).
This dedicated step also lets us audit, log, and ablate state updates without touching the action or recalibration modules.
The discrete extraction satisfies Desideratum v because the action-relevant variables become explicit, schema-validated fields that downstream modules read directly.

\paragraph{Act.}
A monolithic delegate emitting text directly from $\mathbf{m}_t$ and $\mathbf{c}_{u^\star}$ would conflate strategic reasoning (what to say, when, whether at all) with surface generation, leaving no point at which abstention can be a first choice on a specific claim.
We therefore split the action policy into two tiers: a \texttt{Controller} that handles the strategic decision, and a \texttt{Generator} that realizes it as surface text.
The \texttt{Controller} (\Cref{prompt:controller}) reads $\mathbf{m}_t$ and $\mathbf{c}_{u^\star}$ and selects $a_t \in \{\textsc{Silent}, \textsc{Speak}\}$, delegating evaluation to three helpers (\Cref{prompt:helpers}): a candidate curator ranks briefing items by current relevance, a coverage evaluator estimates whether those points remain useful given prior coverage, and a speaker scorer gates by floor dynamics and other participants' topical fit.
On \textsc{Speak}, the \texttt{Controller} commits to a discrete proposition $z_t$ before the \texttt{Generator} (\Cref{prompt:generator}) realizes it in the participant's style, following the plan-then-realize decomposition of classical NLG \citep{reiter2000nlg}.
This commit-before-realize split satisfies Desideratum (iii) as \textsc{Silent} becomes an explicit accept-or-reject on a specific claim, and scoring both actions through the helpers realizes Desideratum iv.

\paragraph{Recalibrate.}
Reconciling the delegate's expectation with the observed outcome requires distinguishing two error sources, a wrong reading of the meeting state and a wrong choice given a correct reading.
A single post-hoc verdict on the delivered action would conflate them, so we combine a withheld forecast with two complementary judges.
Before the delegate acts at $t$, a \texttt{Predictor} (\Cref{prompt:predictor}) issues an intent-level forecast of the next turn from the pre-action $\mathbf{m}_t$ and recent dialogue, hidden from the action policy and reserved for post-hoc evaluation.
Once $\mathbf{H}_{t+1:t+k}$ is observable, an Environment Judge (\Cref{prompt:env-judge}) compares the forecast against the actual continuation, isolating perception error. A Delegate Judge (\Cref{prompt:delegate-judge}) scores the action against the same window, isolating action error.
The \texttt{Recalibrator} (\Cref{prompt:recalibrator}) then fuses both signals into a structured update on $\mathbf{m}_t$ anchored to $\mathbf{o}_{t+1}$, plus guidance for the next decision.
This loop satisfies Desideratum vi and i, since updates only consume turns observed after the action.

\paragraph{Preparation.}

As the participant context $\mathbf{c}_{u^\star}$ serves as the delegate's only static dependency, its construction is decoupled from the runtime loop. 
The long-term profile $\mathbf{p}_{u^\star}$ is extracted from an isolated preceding transcript segment, ensuring strict separation between inference and evaluation data to prevent leakage.
Concurrently, the session briefing $\mathbf{b}_{u^\star}$ synthesizes the participant's objectives and established claims into structured, forward-looking representations that the \texttt{Controller} can directly evaluate. 
Both $\mathbf{p}_{u^\star}$ and $\mathbf{b}_{u^\star}$ are generated via schema-constrained LLMs \cite{kirstein-etal-2024-tell} to guarantee standardized JSON outputs for downstream modules, and constructed only from information available before the interaction segment, thereby satisfying the stability requirement of Desideratum ii.
The briefing constructor does not observe target idea units, cue turns, anchor turns, future participant utterances, or any contribution from the future.

%% file: text_short/04_experimental_setup.tex
\label{sec:experimental_setup}

\subsection{Dataset}
\label{subsec:dataset}

We evaluate on the scenario portion of the AMI Meeting Corpus \citep{carletta2005ami}, in which four participants (project manager, industrial designer, UI specialist, marketing expert) design a remote control over $\sim$30-minute meetings.
We use the full 137-meeting dataset.
Mechanism ablations and robustness checks use \emph{matched} $20$-meeting subsets, sharing meetings, episode set, and briefings so reported $\Delta$s are condition-specific. 
Default settings are $N{=}20$ preceding utterances, evaluation window $k{=}5$.

\subsection{Systems}
\label{subsec:systems}

The main comparison is \sysname{} against two raw-context baselines: \emph{transcript-only} delegate \citep{hu-etal-2025-meeting} reimplemented under our protocol with backbone, participant context, judges, and episode schedule fixed (prompt in \Cref{sec:appendix-prompts});
The \textsc{Reflexion-style} baseline \cite{shinn2023_reflexion} extends the transcript-only delegate with rolling post-action procedural memory, but does not expose structured state fields, helper modules, or state-directed recalibration.
A state-ablated \sysname{} variant keeps the modular pipeline and gets a longer $50$-turn raw window but drops the meeting state and its recalibration.
A third variant keeps the state but disables recalibration (\Cref{sec:ablations-recalib}). 
All systems use GPT-4o \cite{openai2024gpt4} with decoding fixed, and backbone portability uses Gemini-2.5-Pro \cite{comanici2025gemini25} on the matched subset, with settings in \Cref{app:setup}.

\subsection{Evaluation Protocol}
\label{sec:protocol}

We evaluate at the level of participant-owned \emph{idea units}, the proposition-level claims, proposals, or decision-relevant facts the target participant introduces. 
To preclude leakage, each meeting is split.
The first 60\% constructs the participant profile, and the final 40\% (the \emph{interaction segment}) is the evaluation testbed.

\paragraph{Episode construction.}
Each idea unit instantiates a bounded \emph{episode} with two reference points.
The \emph{cue} fixes the information boundary at which the delegate must act, using only causally observed context.
The \emph{anchor} is the ground-truth turn at which the participant articulated the target idea.
The delegate then advances through a $k$-turn evaluation window after the cue \citep{liesenfeld-dingemanse-2024-interactive}, with state updates consuming ground-truth turns only once they are causally observable.

\paragraph{Outcomes.}
A contribution earns recall credit only when (i) it matches the target proposition and (ii) it is neither hallucinated nor redundant.
Each episode resolves into one of four outcomes. \emph{Strict hit}: credited recovery at the anchor ($\Delta=0$). \emph{Loose hit}: credited recovery within the $k$-turn window after the cue ($0 \le \Delta \le k$). \emph{Uncredited attempt}: the agent spoke but the utterance failed at least one credit criterion. \emph{Not-attempted}: silent throughout. This separates \emph{opportunity recognition} (did the agent recover the idea in time?) from \emph{intervention quality} (how well did it intervene?), which we score independently. 
Uncredited attempts stratify into five categories analyzed in \Cref{sec:results-uncredited}.

\paragraph{Metrics and quality.}
\emph{Decision F1} quantifies floor-taking alignment at the anchor.
During speaking turns, output quality is assessed via three binary indicators: \emph{hallucination} (content unsupported by $\mathbf{m}_t$, $\mathbf{c}_{u^\star}$, or the preceding dialogue), \emph{redundancy} (reiteration of previously resolved issues), and \emph{off-topic} (divergence from the active topic). 
Furthermore, LLM-based judges evaluate \emph{grounding}, \emph{relevance}, \emph{decision appropriateness}, and \emph{context appropriateness} (for accepted matches) on a continuous 0--100 scale.
All metrics are macro-averaged at the meeting level, reporting 95\% bootstrap confidence intervals.

\paragraph{Human validation.}
Because LLM judges might be biased \citep{chen-etal-2024-judge-bias}, we calibrate the idea-unit extractor and same-proposition judge against a stratified human-inspection sample.
Three trained annotators (C1+ English) inspected $150$ extracted idea units (balanced across the four participant roles) and $150$ delegate--target same-proposition decisions (balanced across the credited and uncredited outcome categories), with a calibration round before independent scoring and joint adjudication of disagreements.
Aggregate agreement between the adjudicated human label and the LLM judge reaches Cohen's $\kappa = 0.71$.
The full protocol is stated in \Cref{app:human-eval}.

%% file: text_short/05_results.tex
\providecommand{\ci}[1]{{\scriptsize #1}}
\label{sec:results}

This section reports three system-level findings on the $137$-meeting AMI evaluation.

\subsection{Coverage and Output Quality}
\label{sec:results-coverage}

An explicit meeting state lifts prompt-only LLM delegates out of silent abstention without degrading grounding fidelity, an effect that raw-context scaling alone cannot achieve.

\paragraph{Engagement gain.} \sysname{} reduces missed opportunities from the dominant failure mode to a marginal one.
\Cref{tab:headline} compares \sysname{} against two raw-transcript
baselines: a transcript-only delegate, the closest controlled instantiation of \citealp{hu-etal-2025-meeting} under our protocol, and
a lightweight Reflexion-style variant that adds post-action procedural memory but no explicit state, helpers, or state-directed recalibration.
Both retain participant context and receive a longer $50$-turn raw window in place of $\mathbf{m}_t$.
On the $137$-meeting evaluation, \sysname{} doubles credited recovery (loose recall $26.1\!\to\!52.2$, strict $10.7\!\to\!25.1$) and raises Decision F1 by $24.9$ points ($38.1\!\to\!63.0$).
The majority of this gain stems from improved opportunity recognition, isolating the baseline's core deficit to the long-range information dependencies of multi-party meetings.

\begin{table*}[t]
\centering
\scriptsize
\setlength{\tabcolsep}{6pt}
\renewcommand{\arraystretch}{1.15}
\begin{tabular}{lcccccc}
\toprule
\textbf{System} & \makecell{\textbf{Strict}\\\textbf{recall} $\uparrow$} & \makecell{\textbf{Loose}\\\textbf{recall} $\uparrow$} & \makecell{\textbf{Uncredited}\\$\downarrow$} & \makecell{\textbf{Not}\\\textbf{attempted} $\downarrow$} & \makecell{\textbf{Decision}\\\textbf{F1} $\uparrow$} & \makecell{\textbf{Hallucination}\\$\downarrow$} \\
\midrule
Transcript-only & 10.7 & 26.1 & 22.5 & 51.4 & 38.1 & 1.5 \\
Reflexion & 10.3 & 24.3 & 26.1 & 49.5 & 47.6 & 0.3 \\
\rowcolor{ourrow}\textbf{\sysname{} (ours)} & \textbf{25.1} & \textbf{52.2} & 45.3 & \textbf{2.5} & \textbf{63.0} & \textbf{0.6} \\
\bottomrule
\end{tabular}
\caption{Headline comparison on 137 AMI meetings, macro-averaged \%, bold = best per column. Transcript-only gets a 50-turn raw window in place of $\mathbf{m}_t$. Metric definitions in \Cref{sec:protocol}.}
\label{tab:headline}
\end{table*}

\begin{figure}[t]
  \centering
  \includegraphics[width=\columnwidth]{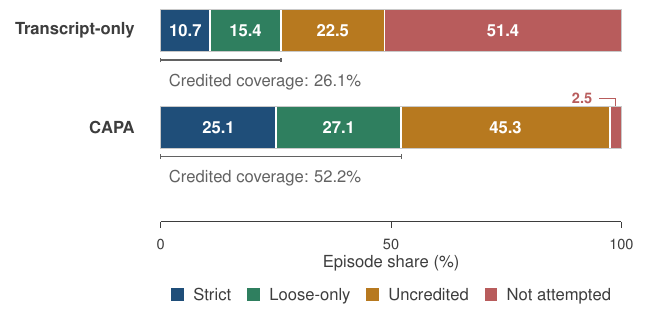}
  \caption{Episode-outcome distribution (\%) over $1{,}126$ idea-unit episodes. Per-system breakdown with $95\%$ bootstrap CIs in \Cref{app:outcome-decomposition}.}
  \label{fig:idea-coverage}
\end{figure}

\paragraph{Outcome redistribution.} The drop in silent abstention translates directly into active, traceable engagement, with the remainder landing as uncredited attempts.
\Cref{fig:idea-coverage} shows the per-system outcome distribution over $1{,}126$ idea-unit episodes.
\sysname{} absorbs the $48.9\%$ drop in not-attempted episodes through $+26.1\%$ credited and $+22.8\%$ uncredited attempts.
Within the credited share, the $27.1\%$ strict--loose gap reflects interventions arriving off-anchor inside the $k{=}5$ window, consistent with backchannels and pacing that postpone articulation. The composition of the uncredited bucket is taken up in \Cref{sec:results-uncredited}.

\paragraph{Grounding preserved.} The coverage gain carries no fabrication or repetition cost.
Hallucination remains at $0.6\%$ and redundancy at $0.0\%$, meaning the active state shifts the agent's participation policy while leaving its evidence grounding intact.
Off-topic rises by $5.2\%$ per episode against $-48.9\%$ not-attempted, the coverage--restraint trade-off that remains an open design question for fixed-weight LLM agents \citep{zhu-etal-2025-grait}.
We read this profile as evidence that the meeting state shifts \emph{participation policy} while leaving evidence grounding unchanged. Component-level attribution is isolated in \Cref{sec:ablations-state} and \Cref{sec:ablations-recalib}.

\subsection{Engagement Timing}
\label{sec:results-timing}
\sysname{} systematically anticipates intervention windows, revealing that the baseline's primary deficit lies in the binary floor-taking decision rather than in its subsequent timing mechanics.

\paragraph{Pre-anchor latency.} 
When \sysname{} takes the floor, its contributions align with or precede the human participant's actual turn.
\Cref{fig:latency-distribution} reports the anchor-relative lead time of every credited match across the 137-meeting evaluation.
Of \sysname{}'s $595$ credited matches, $93.8\%$ are anchor-aligned or pre-anchor, with a median of $1.0$ turn before the anchor (under the convention $\ge 0$ synchronized, $>0$ pre-anchor).
This timing precision drives the absolute $24.9\%$ increase in Decision F1 over the baseline ($38.1\!\to\!63.0$).

\paragraph{The floor-taking bottleneck.} Comparing both systems shows they diverge in how often they engage, but share identical timing distributions when they do.
Intervention rates differ by $48.9\%$ ($97.5\%$ for \sysname{} versus $48.6\%$ for the baseline), while the baseline's lead-time distribution on its few credited matches mirrors \sysname{}'s.
This asymmetry isolates the prompt-only failure specifically to opportunity recognition, consistent with prior reports that multi-party turn anticipation is a discrete bottleneck for standard LLMs \citep{penzo-etal-2024-llms, hilgert2025_nextspeaker}.

\input{figures/05_latency_histogram}

\subsection{Uncredited-Attempt Taxonomy}
\label{sec:results-uncredited}
By explicitly tracking meeting state, \sysname{} shifts the baseline's tendency to remain silently inactive with traceable, active engagement. When the system misses a target, it produces diagnosable near-misses rather than opaque omissions.

\paragraph{Error composition.} The $45.3\%$ uncredited attempt rate decomposes into four mutually exclusive categories of selection and timing discrepancies (\Cref{fig:uncredited-taxonomy}).
We classify all $504$ uncredited episodes \sysname{} produces across the $137$-meeting evaluation as: \emph{incomplete match} (correct proposition, missing qualifier), \emph{different proposition} (a briefing-supported point distinct from the reference), \emph{late match} (correct proposition generated outside the $k$-turn window), and \emph{safety/quality flag} (off-topic or redundancy filter triggered).
This error taxonomy maps to the system's modular components.
\emph{Incomplete match} isolates a realization deficit in the \texttt{Generator}. 
\emph{Different proposition} errors indicate imprecise candidate curation against. 
\emph{Late matches} reflect timing misalignment driven by the $\mathbf{m}_t$ and the speaker scorer.
Finally, \emph{safety/quality flags} denote expected candidate rejections by the output filter during drafting.

\input{figures/04_uncredited_taxonomy}

\paragraph{System observability.} This shift from omission to selection yields a methodological benefit for system diagnostics.
The transcript-only delegate's $51.4\%$ silent-abstention rate produces no decision trace, precluding error attribution.
\sysname{}'s uncredited attempts generate a stepwise record, isolating failures and enabling targeted refinement.

%% file: figures/05_latency_histogram.tex
% \begin{figure}[t]
% \centering
% \setlength{\fboxsep}{4pt}
% \fbox{\parbox{0.95\columnwidth}{\footnotesize\raggedright
% \textcolor{red}{\textbf{[Latency-histogram figure --- to be filled.]}
% Required content: histogram (or compact violin) of the credited-match
% latency distribution for \sysname{} over the $137$-meeting evaluation, in
% units of turns (${\geq}0$ = synchronised, ${>}0$ = before anchor, ${<}0$
% = after anchor). Suggested $x$-axis range $[-3, +5]$ turns. Required
% data: per-episode latency for every credited match (strict or
% loose-only); easy dump from existing main-run logs, no new runs.}

% \textcolor{red}{\textbf{Expected interpretation (to be confirmed once
% plotted):} the distribution is non-negative-biased with median $1.0$;
% the median-before-anchor pattern is the visual case that \sysname{} is
% \emph{attempting} anticipation rather than late recovery. Also useful:
% if Hu-style transcript-only credited matches can be similarly plotted,
% show both as overlapping histograms.}
% }}
% \caption{\textcolor{red}{\textbf{Credited-match latency distribution (to
% be filled).} Visualises the $1.0$-turn median latency finding from
% \S\ref{sec:results-rq1}: \sysname{}'s credited contributions cluster
% \emph{before} the participant's anchor turn rather than after. Priority:
% MEDIUM-HIGH -- the figure converts a single median number into a
% behavioural pattern the reader sees.}}
% \label{fig:latency-distribution}
% \end{figure}

\begin{figure}[t]
\centering
\includegraphics[width=\columnwidth]{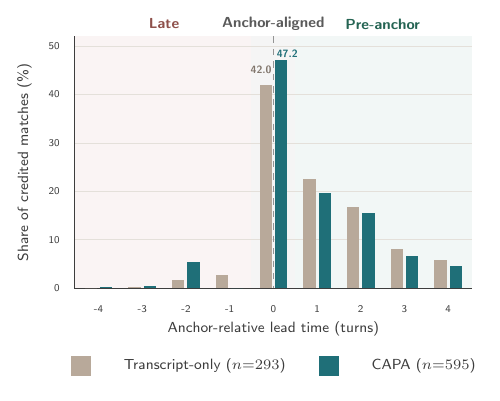}
\caption{Timing distribution of strict and loose matches for \sysname{} (n=595) and the transcript-only baseline (n=293). Positive values indicate pre-anchor contributions, 0 represents anchor-aligned matches, and negative values indicate late recovery.}
\label{fig:latency-distribution}
\end{figure}

%% file: figures/04_uncredited_taxonomy.tex
\begin{figure}[t]
  \centering
  \includegraphics[width=\columnwidth]{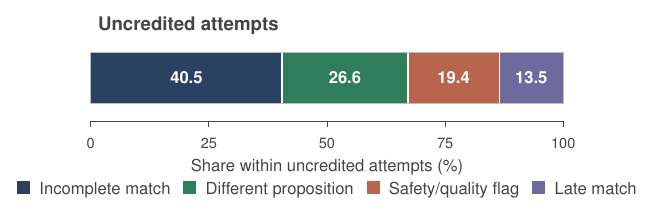}
  \caption{Uncredited-attempt taxonomy over all $504$ uncredited episodes \sysname{} produces across the $137$-meeting evaluation ($45.3\%$ uncredited bucket). Each attempt is assigned to exactly one of four mutually exclusive failure types defined in \Cref{sec:results-uncredited}.}
  \label{fig:uncredited-taxonomy}
\end{figure}

%% file: text_short/06_ablations.tex
\label{sec:ablations}
To understand the mechanistic drivers behind \sysname{}'s performance, we isolate its explicit state tracking and post-action recalibration.
Because the architecture decouples the floor-taking decision from generation, we can independently ablate these modules to trace their specific effects on agent behavior.
By removing these components, we show that state tracking exclusively governs the binary participation policy, while recalibration refines the qualitative selection of what is said. 
Finally, we verify robustness across varying context and evaluation windows, underlying LLM backbones, and a non-scenario meeting corpus.
Comprehensive ablation tables and experiment descriptions are provided in \Cref{app:robustness-tables}.

\subsection{Meeting State Carries the Threshold}
\label{sec:ablations-state}

The explicit meeting state closes the recognition gap in multi-party delegation.
On a 20-meeting subset, we substitute $\mathbf{m}_t$ and the recalibration step for a longer $50$-turn raw window (\Cref{tab:state-ablation}). 
This leads to a loose recall drop by $22.4\%$, strict recall reduces by $14.5\%$ and Decision F1 by $24.7\%$, while redundancy increases by $10.7\%$.
The reduction in behavioral metrics against a static grounding score confirms that the state mechanism determines whether and when to engage.
\Cref{fig:qualitative-main} makes this mechanism concrete as the state exposes an open question, participant fit, and an uncovered briefing point that the raw-transcript delegate does not act on.
\input{figures/03_qualitative_example_main}

\subsection{Recalibration Refines Selection}
\label{sec:ablations-recalib}
Post-action recalibration narrows contribution selection without altering the floor-taking threshold, demonstrating that state-update feedback and output-rewriting feedback target distinct failure modes (\Cref{tab:recalibration}).
Removing only the recalibration step while keeping $\mathbf{m}_t$ and the upstream stack intact causes loose recall to drop by $3.9\%$ and uncredited attempts to rise by $4.6\%$.
Meanwhile, Decision F1 and grounding remain statistically unchanged.
This confirms that output-rewriting baselines (e.g., Reflexion) are structurally inapplicable to our primary failure regime: they exist to refine drafted utterances, whereas multi-party delegates primarily fail via silent abstention, leaving no draft to refine.

\subsection{Backbone Model Portability} % and Safety--Coverage Calibration
\label{sec:ablations-backbone}

The core architectural mechanism transfers across backbones, though the specific safety--coverage operating point is model-dependent.
Across GPT-4o, Gemini-2.5-Pro, Llama-3.3-70B, and Qwen3.6-27B, Decision F1 stays within $57.9$ to $69.2$ while coverage and precision trade off along one curve.
Permissive backbones such as Gemini gain recall at the cost of off-topic contributions. 
Conservative backbones such as Qwen abstain more ($17.8\%$, still about a third of the transcript-only baseline's $51.4\%$) while selecting most cleanly (off-topic $3.3\%$, hallucination $0.0\%$).
A new backbone can be calibrated on held-out episodes by tightening the printed suitability and restraint constants until precision meets the deployment requirement, with no retraining (full table in \Cref{tab:model-comparison}).

\subsection{Transfer Beyond AMI}
\label{sec:results-icsi}
As a cross-corpus robustness check, we test whether \sysname{} depends on AMI's scripted four-person format. We replay 10 ICSI meetings, i.e., real research-group discussions with six participants on average and up to ten \citep{janin2003icsi}, under the identical protocol, adapting only the profile construction.
We observe a similar behavior, where silence remains near zero ($1.2\%$ versus $2.5\%$ on AMI), loose recall is higher ($73.6\%$ [61.1, 86.0] versus $52.2\%$), Decision F1 is comparable ($64.8\%$ [57.4, 74.9] versus $63.0\%$), and hallucination stays low ($1.8\%$), with redundancy at $0.0\%$ and off-topic at $4.9\%$. The full table reports 42 participant runs, 47 episodes, and 235 evaluated steps (\Cref{tab:icsi-transfer}).

\subsection{Sensitivity to Evaluation Parameters}
\label{sec:ablations-sensitivity}

The primary findings remain stable under variations in the raw-window length and exhibit expected coverage--horizon trade-offs.
\textbf{Raw context window (N).} With $\mathbf{m}_t$ available, the raw transcript window is no longer load-bearing.
Across $N \in \{10, 20, 30\}$, loose recall remains stable between $50.5\%$ and $56.7\%$, and Decision F1 between $61.1\%$ and $63.1\%$ (\Cref{app:context-window}).
\textbf{Evaluation window ($k$).} Expanding $k$ admits more credited matches (loose recall rises from $43.9\!\to\!62.7$ across $k \in \{3, 5, 7, 9\}$) without artificially inflating the system's anchor synchrony (\Cref{app:window}).

%% file: figures/03_qualitative_example_main.tex
\begin{figure}[t]
\centering
\scriptsize
\begin{tcolorbox}[
  colback=gray!2,
  colframe=gray!45,
  boxrule=0.35pt,
  arc=1pt,
  left=2pt,right=2pt,top=2pt,bottom=2pt,
  width=\columnwidth,
  boxsep=1pt
]
\textbf{Cue.} The group asks, ``Are LEDs beneath the buttons?'' The UI participant has an uncovered briefing point about LED placement.\\[-1pt]
\textbf{State read.} Open question; UI has highest speaker fit; LED placement remains uncovered.\\[-1pt]
\textbf{Transcript-only.} \textsc{Silent}.\\[-1pt]
\textbf{CAPA.} \textsc{Speak}: ``One idea could be to integrate the LEDs around the buttons or even within them.''\\[-1pt]
\textbf{Reference.} The UI participant next says, ``Yeah, around the buttons, or in the buttons even.'' $\Rightarrow$ strict match, no safety flag.
\end{tcolorbox}
\caption{Compact qualitative trace. Explicit floor, speaker-fit, and coverage fields support a timely contribution where the raw-transcript delegate remains silent. The complete trace is in \Cref{app:qualitative-diagnostics}.}
\label{fig:qualitative-main}
\end{figure}

%% file: text_short/06_conclusion.tex
In online meeting delegation, prompt-only LLM proxies fail to recognize when to intervene, defaulting to silent abstention during most talking opportunities.
We introduced \sysname{}, a \emph{perceive--act--recalibrate} loop operating over an explicit meeting state, as an architectural response to this recognition gap.
To evaluate it, we designed an episode-level protocol that scores whether, when, and what a delegate contributes relative to a human participant's actual idea units.
On 137 AMI meetings, \sysname{} reduced the silence rate from $51.4\%$ to $2.5\%$, doubled credited recovery ($26.1\!\to\!52.2$), and preserved grounding fidelity with a $0.6\%$ hallucination rate.
Our findings demonstrate that effective multi-party recognition requires maintaining variables like stance, coverage, and floor control as explicit state fields, as LLMs cannot reliably recover them from raw context streams.
By decoupling the floor-taking decision from surface generation, \sysname{} resolves the upstream recognition error and allows for transparent, module-attributable selection errors.
Furthermore, we establish that this state-updating correction channel is non-substitutable by output-rewriting paradigms as they are poorly matched to the primary failure regime we observe.
Robustness checks confirm the core architecture transfers across configurations.
The generalization of this state-driven architecture to other domains with similar causal information boundaries, e.g., customer-support handoff, remain future directions.

%% file: text_short/07_limitations.tex
The primary evaluation of \sysname{} focuses on the AMI scenario portion. Alternative corpora do not provide all protocol conditions without adaptation. ELITR \citep{nedoluzhko-etal-2022-elitr}, used by related meeting delegation work \citep{hu-etal-2025-meeting}, and ICSI \citep{janin2003icsi} cover conceptually similar formal settings but lack the continuity that AMI provides. For the ICSI robustness probe (\Cref{sec:results-icsi}), we therefore adapt only profile construction while holding causal replay fixed.
Other meeting datasets such as parliamentary or formal institutional datasets \citep{hu-etal-2023-meetingbank,zhong-etal-2021-qmsum} enforce rigid turn-taking structures where spontaneous floor-taking is structurally not possible.

The system's performance depends significantly on the capabilities of the underlying language model.
While our implementation uses GPT-4o, models with different reasoning capabilities or smaller context windows may produce less accurate delegation capabilities.
Our ablation studies suggest that performance can carry over to other model families, demonstrating the framework's architectural robustness.

%% file: text_short/08_ethics.tex
\sysname{} is evaluated offline on the publicly released AMI Meeting
Corpus (CC BY 4.0). AMI participants consented to recording for
research use, which we treat as a sufficient basis for offline
transcript analysis but not for live impersonation. Any deployment of
state-conditioned delegation in a live meeting would require explicit,
recurring consent from all participants present, because the
architecture lets the agent surface a participant-owned proposition the
represented participant may not have stated themselves. Misuse paths
include impersonation under false pretenses, plausible-deniability
speech laundering, and asymmetric advantage of represented over
unrepresented participants. We release \sysname{} under MIT license as
research infrastructure. Production use should go through participant
consent.

%% file: text_short/B1_resources.tex
\section{Open Resources and Licensing}
\label{app:resources}

\subsection{Repository and License}
\label{app:repository}
The code for this work is available on \url{https://github.com/FKIRSTE/emnlp2026-meeting-delegation} under an MIT license together with the evaluation protocol, prompt templates, judges, and analysis scripts after acceptance.

\subsection{Datasets and Licensing}
We evaluate on the scenario portion of the AMI Meeting Corpus \citep{carletta2005ami}, distributed under CC BY 4.0. \Cref{tab:datasets} summarizes the license and the high-level corpus statistics.

\begin{table}[h]
\centering
\footnotesize
\setlength{\tabcolsep}{3.5pt}
\renewcommand{\arraystretch}{1.1}
\begin{tabularx}{\columnwidth}{@{}llrrX@{}}
\toprule
\textbf{Dataset} & \textbf{License} & \textbf{Size} & \textbf{Domain} \\
\midrule
AMI (scenario) & CC BY 4.0 & 137 &
remote-control design \\
\bottomrule
\end{tabularx}
\caption{Dataset licensing and overview. \emph{Size}: number
of meetings used.}
\label{tab:datasets}
\end{table}

%% file: text_short/B2_setup.tex
\section{Experimental Setup Details}
\label{app:setup}

\subsection{Implementation}
We implement \sysname{} in Python. All LLM calls go through a single API wrapper with retry-on-error and structured JSON output validation against the schema for each module (\texttt{Perceiver}, \texttt{Predictor}, Environment Judge, Delegate Judge, \texttt{Recalibrator}, \texttt{Controller}, \texttt{Generator}, helpers). Schema-constrained decoding guarantees that every module emits parseable structured output. The \texttt{Shared Memory} is a typed dictionary indexed by participant, meeting, and turn.

\subsection{Model Specs}
\label{app:model-specs}
GPT-4o is the default backbone for all main experiments. The backbone-portability study evaluates Gemini-2.5-Pro, Llama-3.3-70B, and Qwen3.6-27B. \Cref{tab:models} lists the reported endpoints; decoding settings remain fixed.

\begin{table}[h]
\centering
\footnotesize
\setlength{\tabcolsep}{3pt}
\renewcommand{\arraystretch}{1.15}
\begin{adjustbox}{max width=\columnwidth}
\begin{tabular}{l|ccc}
\toprule
\textbf{Model} & \textbf{Snapshot} & \textbf{Parameters} & \textbf{Provider} \\
\midrule
GPT-4o          & {2024-11-20} & $\sim 200$B (est.) & OpenAI \\
Gemini-2.5-Pro  & {2025-06-17} & not disclosed      & Google \\
Llama-3.3-70B & {2024-12-06} & 70B & Groq \\
Qwen3.6-27B   & {2026-04-21} & 27B & Groq \\
\bottomrule
\end{tabular}
\end{adjustbox}
\caption{Model snapshots and providers. GPT-4o is the default backbone. Snapshot names come from provider metadata when exposed.}
\label{tab:models}
\end{table}

\subsection{Hyperparameters}
\label{app:hyperparameters}
We keep default values for top-$p$ ($1.0$), frequency penalty ($0.0$), and presence penalty ($0.0$). Temperature is $T{=}0.1$ for every LLM module except the \texttt{Generator}, which runs at $T{=}0.3$ to allow modest natural-language variation after the \texttt{Controller} has committed to a semantic point. Decoding parameters are held fixed across systems and conditions. Default episode parameters are $N{=}20$ preceding utterances and evaluation window $k{=}5$, both varied in the robustness checks (\Cref{app:robustness}). All metrics are macro-averaged at the meeting level with $95\%$ bootstrap CIs ($B{=}10{,}000$).

%% file: text_short/A2_compute.tex
\subsection{Computational Footprint}
\label{app:compute}

We distinguish response latency from total computation per decision. Only the \texttt{Controller}, its three helpers, and the conditional \texttt{Generator} lie on the response path. The \texttt{Perceiver} runs once per episode and can overlap with ongoing speech. The \texttt{Predictor}, judges, and \texttt{Recalibrator} execute after the floor decision and affect later turns. The critical-path rows therefore measure the delay before a contribution, while the full-decision rows measure all computation attributed to one decision. All reported latencies are wall-clock measurements from the experimental harness and include its fixed three-second post-call throttle.

{\color{blue}
\begin{table*}[]
\centering
\small
\setlength{\tabcolsep}{2.5pt}
\renewcommand{\arraystretch}{1.08}
\begin{adjustbox}{max width=\textwidth}
\begin{tabular}{lrlrrrrrr}
\toprule
\textbf{Module} & \textbf{Calls/dec.} & \textbf{Timing} & \textbf{Mean s} & \textbf{p95 s} & \textbf{Input tok.} & \textbf{Cached input} & \textbf{Output tok.} & \textbf{Cost/dec.} \\
\midrule
Perceiver$^\dagger$ & 0.845 & pre-episode & 6.27 & 23.16 & 10,322 & 5,272 & 787 & \$0.0271 \\
Predictor & 1.000 & post-turn & 3.90 & 4.08 & 4,540 & 1,439 & 54 & \$0.0101 \\
Controller + 3 helpers & 2.991 & floor path & 19.85 & 33.51 & 46,009 & 26,195 & 2,247 & \$0.1048 \\
Generator & 0.336 & floor path, conditional & 1.28 & 3.91 & 1,242 & 904 & 16 & \$0.0021 \\
Environment Judge & 1.000 & post-turn & 4.14 & 4.37 & 7,198 & 5,120 & 98 & \$0.0126 \\
Delegate Judge & 1.345 & post-turn & 6.50 & 10.81 & 10,619 & 5,750 & 317 & \$0.0225 \\
Recalibrator & 2.000 & post-turn & 10.33 & 11.19 & 10,658 & 5,632 & 1,025 & \$0.0299 \\
\midrule
\textbf{Critical path (floor-taking)} & \textbf{3.327} & & \textbf{21.14} & \textbf{36.49} & \textbf{47,251} & \textbf{27,099} & \textbf{2,263} & \textbf{\$0.1069} \\
\textbf{Runtime per decision, excluding Perceiver} & \textbf{8.673} & floor + post-turn & \textbf{46.03} & \textbf{65.12} & \textbf{80,267} & \textbf{45,040} & \textbf{3,757} & \textbf{\$0.1819} \\
\textbf{Architecture per decision, including Perceiver} & \textbf{9.518} & amortized & \textbf{52.30}$^\dagger$ & -- & \textbf{90,590} & \textbf{50,312} & \textbf{4,544} & \textbf{\$0.2090} \\
\bottomrule
\end{tabular}
\end{adjustbox}
\caption{GPT-4o latency, token use, and cost over 110 decisions in 22 five-step episodes. CAPA spoke on 37 decisions; the floor-taking LLM path ran on 72.}
\label{tab:compute-gpt}
\end{table*}
}

\begin{table*}[]
\centering
\small
\setlength{\tabcolsep}{3pt}
\renewcommand{\arraystretch}{1.08}
\begin{adjustbox}{max width=\textwidth}
\begin{tabular}{lrlrrrrr}
\toprule
\textbf{Module} & \textbf{Calls/dec.} & \textbf{Timing} & \textbf{Mean s} & \textbf{p95 s} & \textbf{Input tok./dec.} & \textbf{Output tok./dec.} & \textbf{Cost/dec.} \\
\midrule
Perceiver$^{\dagger\ddagger}$ & 0.714 & pre-episode & 5.32 & 12.21 & $\approx$11,617 & $\approx$753 & $\approx$\$0.00923 \\
Predictor & 1.000 & post-turn & 3.59 & 3.65 & 4,197 & 64 & \$0.00271 \\
Controller + 3 helpers & 2.842 & floor path & 18.35 & 31.97 & 44,922 & 2,982 & \$0.03590 \\
Generator & 0.189 & floor path, conditional & 0.67 & 3.53 & 671 & 9 & \$0.00043 \\
Environment Judge & 1.000 & post-turn & 4.04 & 4.19 & 6,903 & 191 & \$0.00472 \\
Delegate Judge & 1.189 & post-turn & 4.76 & 7.81 & 8,296 & 200 & \$0.00558 \\
Recalibrator & 2.000 & post-turn & 9.10 & 9.64 & 10,757 & 1,037 & \$0.00956 \\
\midrule
\textbf{Critical path (floor-taking)} & \textbf{3.032} & & \textbf{19.03} & \textbf{34.86} & \textbf{45,592} & \textbf{2,991} & \textbf{\$0.03633} \\
\textbf{Runtime per decision, excluding Perceiver} & \textbf{8.221} & floor + post-turn & \textbf{40.54} & \textbf{59.39} & \textbf{75,746} & \textbf{4,483} & \textbf{\$0.05890} \\
\textbf{Architecture per decision, including Perceiver} & \textbf{8.935} & amortized & \textbf{45.85}$^\dagger$ & -- & $\approx$\textbf{87,363} & $\approx$\textbf{5,235} & $\approx$\textbf{\$0.06813} \\
\bottomrule
\end{tabular}
\end{adjustbox}
\caption{Qwen3.6-27B computational footprint over 95 decisions in 19 complete episodes.}
\label{tab:compute-qwen}
\end{table*}

\begin{table*}[]
\centering
\small
\setlength{\tabcolsep}{3pt}
\renewcommand{\arraystretch}{1.08}
\begin{adjustbox}{max width=\textwidth}
\begin{tabular}{lrlrrrrr}
\toprule
\textbf{Module} & \textbf{Calls/dec.} & \textbf{Timing} & \textbf{Mean s} & \textbf{p95 s} & \textbf{Input tok./dec.} & \textbf{Output tok./dec.} & \textbf{Cost/dec.} \\
\midrule
Perceiver$^{\dagger\ddagger}$ & 0.724 & pre-episode & 5.70 & 22.38 & $\approx$9,011 & $\approx$656 & $\approx$\$0.00584 \\
Predictor & 1.000 & post-turn & 3.49 & 3.75 & 3,505 & 55 & \$0.00211 \\
Controller + 3 helpers & 2.781 & floor path & 15.80 & 27.69 & 38,353 & 1,743 & \$0.02401 \\
Generator & 0.448 & floor path, conditional & 1.58 & 3.67 & 1,525 & 23 & \$0.00092 \\
Environment Judge & 1.000 & post-turn & 3.89 & 4.08 & 6,410 & 120 & \$0.00388 \\
Delegate Judge & 1.448 & post-turn & 5.55 & 7.62 & 8,274 & 135 & \$0.00499 \\
Recalibrator & 2.000 & post-turn & 8.82 & 9.32 & 9,547 & 931 & \$0.00637 \\
\midrule
\textbf{Critical path (floor-taking)} & \textbf{3.229} & & \textbf{17.26} & \textbf{28.04} & \textbf{39,878} & \textbf{1,766} & \textbf{\$0.02492} \\
\textbf{Runtime per decision, excluding Perceiver} & \textbf{8.676} & floor + post-turn & \textbf{39.03} & \textbf{51.32} & \textbf{67,614} & \textbf{3,006} & \textbf{\$0.04227} \\
\textbf{Architecture per decision, including Perceiver} & \textbf{9.400} & amortized & \textbf{44.73}$^\dagger$ & -- & $\approx$\textbf{76,625} & $\approx$\textbf{3,663} & $\approx$\textbf{\$0.04811} \\
\bottomrule
\end{tabular}
\end{adjustbox}
\caption{Llama-3.3-70B computational footprint over 105 decisions in 21 complete episodes.}
\label{tab:compute-llama}
\end{table*}

\paragraph{GPT-4o latency.}
The observed response path takes 21.14 s on average, with a p95 of 36.49 s. The harness throttle contributes 10.01 s to the mean. Removing it analytically gives a provider-only response latency of 11.12 s mean and 19.05 s p95 across all 110 decisions. Among the 72 decisions that execute the floor-taking LLM path, provider calls take 16.99 s mean and 19.39 s p95.

$^\dagger$The \texttt{Perceiver} runs once per five-decision episode. Its episode-level latency is 31.35 s mean and 115.78 s p95, which \Cref{tab:compute-gpt} amortizes across the five decisions. The 52.30 s full-decision mean also includes post-turn computation and therefore does not represent response latency.

\paragraph{Open-weight latency.}
On Groq, Qwen's observed response path takes 19.03 s on average and Llama's takes 17.26 s. Their full-decision means are 45.85 s and 44.73 s because these totals include background updates and the amortized \texttt{Perceiver}. $^\dagger$For both backbones, the \texttt{Perceiver} runs once per five-decision episode and its measured latency is amortized across those decisions.

\paragraph{Cost accounting.}
The GPT-4o response path costs \$0.1069 per decision, and the complete architecture costs \$0.2090 after including background updates and the amortized \texttt{Perceiver}. GPT-4o costs use provider-reported cached-token counts and rates of \$2.50/M uncached input tokens, \$1.25/M cached input tokens, and \$10/M output tokens. For Qwen, the rates are \$0.60/M input tokens and \$3.00/M output tokens. For Llama, they are \$0.59/M input tokens and \$0.79/M output tokens. $^\ddagger$Only open-weight \texttt{Perceiver} token use is estimated; all runtime-module usage is provider-reported.

%% file: text_short/B3_human_eval.tex
\section{Human Evaluation Protocol}
\label{app:human-eval}

\subsection{Annotators}
Three trained annotators (C1+ English) ran the inspection, recruited as research assistants or doctoral candidates with academic background in multi-party dialogue. Each annotator gave explicit consent for the anonymized annotations to be used in this work. The protocol received approval from our institution's ethics committee before any data collection.

\subsection{Sampling and Procedure}
We drew a stratified sample of $150$ extracted idea units, balanced across the four participant roles (\textsc{PM}, \textsc{ID}, \textsc{UI}, \textsc{ME}), and $150$ delegate--target same-proposition decisions, balanced across the credited (strict, loose-only) and uncredited outcome categories. Each idea unit got a binary label (is this a proposition-level idea unit?). Each delegate--target pair got a binary label (does the delegate's utterance express the same proposition as the target?). Annotators were blind to which delegate produced each utterance and to which condition the episode belonged. Each annotator scored independently after a calibration round, and disagreements were adjudicated through joint discussion.

\subsection{Agreement}
We compute Cohen's $\kappa$ between the adjudicated human label and the LLM judge across the judged dimensions (idea-unit extraction, same-proposition matching, hallucination, redundancy, off-topic). Aggregate agreement reaches $\kappa = 0.71$, in the substantial-agreement range.

%% file: text_short/A3_robustness.tex
\section{Mechanism Ablation Tables}
\label{app:robustness-tables}

\begin{figure*}[!t]
  \centering
  \includegraphics[width=.75\linewidth]{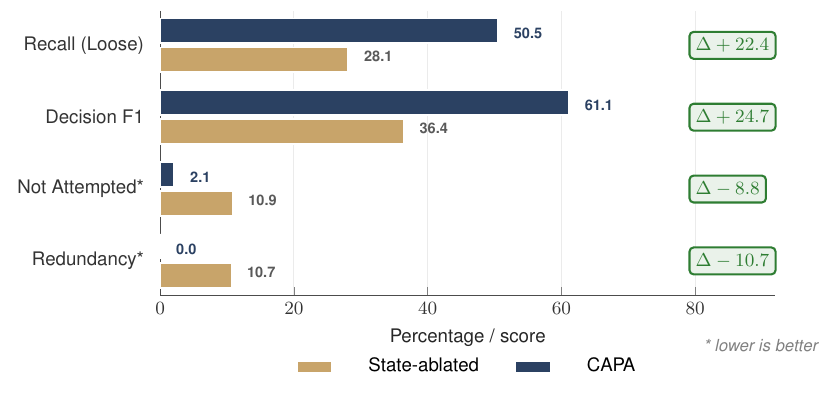}
  \caption{Meeting-state ablation, grouped bar view of \Cref{tab:state-ablation}.}
  \label{fig:state-ablation}
\end{figure*}

\begin{figure*}[!t]
  \centering
  \includegraphics[width=.75\linewidth]{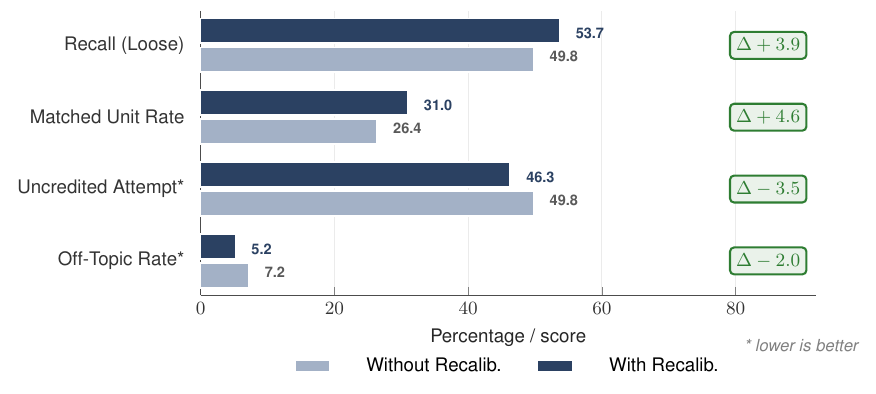}
  \caption{Recalibration ablation, grouped bar view of \Cref{tab:recalibration}.}
  \label{fig:recalibration}
\end{figure*}

The two ablations isolate complementary effects. Removing the meeting state $\mathbf{m}_t$ collapses coverage, Decision F1, and redundancy together (\Cref{tab:state-ablation}). Removing the \texttt{Recalibrator} narrows contribution selection while the floor-taking threshold stays inside bootstrap noise (\Cref{tab:recalibration}). \Cref{fig:state-ablation} and \Cref{fig:recalibration} render both tables as grouped bar charts.

\begin{table}[!t]
\centering
\footnotesize
\setlength{\tabcolsep}{3pt}
\renewcommand{\arraystretch}{1.15}
\begin{adjustbox}{max width=\columnwidth}
\begin{tabular}{l|cc|c}
\toprule
\textbf{Metric} & \textbf{\sysname{} (ours)} & \textbf{State-Ablated} & \textbf{$\Delta$} \\
\midrule
Strict recall (\%) $\uparrow$ & \estci{\textbf{23.0}}{15.4, 30.7} & \estci{8.5}{4.1, 13.4} & $+14.5$ \\
Loose recall (\%) $\uparrow$  & \estci{\textbf{50.5}}{40.7, 59.7} & \estci{28.1}{20.1, 36.8} & $+22.4$ \\
Not Attempted (\%) $\downarrow$ & \estci{\textbf{2.1}}{0.0, 5.9}  & \estci{10.9}{5.9, 16.2}  & $-8.8$  \\
Decision F1 (\%) $\uparrow$   & \estci{\textbf{61.1}}{56.9, 65.1} & \estci{36.4}{33.2, 39.9} & $+24.7$ \\
Redundancy (\%) $\downarrow$  & \estci{\textbf{0.0}}{0.0, 0.0}    & \estci{10.7}{6.7, 15.2}  & $-10.7$ \\
Grounding (Score) $\uparrow$  & \estci{\textbf{74.3}}{73.2, 75.3} & \estci{72.7}{71.7, 73.7} & $+1.6$  \\
\bottomrule
\end{tabular}
\end{adjustbox}
\caption{Meeting-state ablation on 20 matched meetings ($k{=}5$, $B{=}10{,}000$). Bold = better per row. The ablated system omits $\mathbf{m}_t$ and recalibration and gets a 50-turn raw window.}
\label{tab:state-ablation}
\end{table}

\begin{table}[!t]
\centering
\footnotesize
\setlength{\tabcolsep}{3pt}
\renewcommand{\arraystretch}{1.15}
\begin{adjustbox}{max width=\columnwidth}
\begin{tabular}{l|cc}
\toprule
\textbf{Metric} & \textbf{With Recalib.\ (ours)} & \textbf{Without} \\
\midrule
Strict recall (\%) $\uparrow$        & \estci{\textbf{28.0}}{20.7, 36.0} & \estci{26.0}{19.1, 32.8} \\
Loose recall (\%) $\uparrow$         & \estci{\textbf{53.7}}{44.1, 63.6} & \estci{49.8}{41.7, 57.2} \\
Matched-unit rate (\%) $\uparrow$    & \estci{\textbf{31.0}}{23.1, 40.1} & \estci{26.4}{21.0, 32.0} \\
Uncredited attempt (\%) $\downarrow$ & \estci{\textbf{46.3}}{36.4, 55.9} & \estci{49.8}{42.5, 57.9} \\
Decision F1 (\%) $\uparrow$          & \estci{66.2}{61.9, 70.4}          & \estci{\textbf{66.7}}{62.3, 71.0} \\
Off-topic rate (\%) $\downarrow$     & \estci{\textbf{5.2}}{2.3, 8.5}    & \estci{7.2}{3.6, 11.4} \\
Grounding (Score) $\uparrow$         & \estci{74.1}{72.8, 75.3}          & \estci{74.3}{73.1, 75.4} \\
\bottomrule
\end{tabular}
\end{adjustbox}
\caption{Recalibration ablation on 20 matched meetings ($k{=}5$, $B{=}10{,}000$). Bold = better per row. Both variants share $\mathbf{m}_t$, the \texttt{Predictor}, and the judges.}
\label{tab:recalibration}
\end{table}

\section{Conditional Quality Scores}
\label{app:quality-decomp}

\Cref{tab:quality} reports \sysname{}'s judge-based quality scores on the $137$-meeting evaluation. Each dimension is scored only over the steps where it applies. Decision appropriateness covers every decision step, relevance and grounding cover speaking turns, and context appropriateness covers accepted same-proposition matches only. A like-for-like cross-system comparison on these scores is not meaningful here, because the transcript-only delegate is silent in $51.4\%$ of episodes and its conditional distribution is taken over a smaller and more selective subset. We therefore read the absolute numbers as architectural characterization of \sysname{} and defer the within-architecture comparison to the grounding-invariance result under the meeting-state ablation (\Cref{sec:ablations-state}).

\begin{table}[!t]
\centering
\small
\setlength{\tabcolsep}{3pt}
\renewcommand{\arraystretch}{1.15}
\begin{tabular}{l|cc}
\toprule
\textbf{Dimension} & \textbf{Mean} & \textbf{95\% CI} \\
\midrule
Decision appropriateness          & 88.8 & [88.5, 89.0] \\
Context appropriateness$^\dagger$ & 88.0 & [87.4, 88.5] \\
Relevance                         & 78.0 & [77.3, 78.6] \\
Grounding                         & 73.8 & [73.3, 74.3] \\
\bottomrule
\end{tabular}
\caption{\sysname{} judge-based quality scores (0--100, higher better). $\dagger$ scored only on accepted same-proposition matches. $B{=}10{,}000$ bootstrap resamples.}
\label{tab:quality}
\end{table}

\section{Detailed Qualitative Trace}
\label{app:qualitative-diagnostics}

\Cref{fig:qualitative} expands the compact main-paper example with the full observed context, state snapshot, semantic plan, realized utterance, and verifier-backed credit outcome.

\input{figures/03_qualitative_example}

\newpage

\section{Outcome Decomposition}
\label{app:outcome-decomposition}

\Cref{tab:outcome-decomposition} gives the per-system breakdown of the episode outcomes summarized in \Cref{fig:idea-coverage}, with 95\% bootstrap CIs on every rate. The headline shift is the not-attempted rate, which drops from $51.4\%$ for the transcript-only delegate to $2.5\%$ for \sysname{}. The two systems' CIs do not overlap on any outcome category.

\begin{table}[!t]
\centering
\footnotesize
\setlength{\tabcolsep}{3pt}
\renewcommand{\arraystretch}{1.15}
\begin{adjustbox}{max width=\columnwidth}
\begin{tabular}{l|cc}
\toprule
\textbf{Outcome} & \textbf{Transcript-only} & \textbf{\sysname{} (ours)} \\
\midrule
Strict hit         & \estci{10.7}{8.7, 12.8}  & \estci{25.1}{22.1, 28.2} \\
Loose-only hit     & \estci{15.4}{12.7, 18.2} & \estci{27.1}{24.3, 29.9} \\
Uncredited attempt & \estci{22.5}{19.7, 25.3} & \estci{45.3}{41.7, 48.9} \\
Not attempted      & \estci{51.4}{47.8, 55.2} & \estci{2.5}{1.6, 3.7} \\
\bottomrule
\end{tabular}
\end{adjustbox}
\caption{Per-system outcome decomposition over the $137$-meeting fixed-window episode set. Rates are macro-averaged over meetings with retained official episodes ($B{=}10{,}000$ bootstrap resamples).}
\label{tab:outcome-decomposition}
\end{table}

\section{ICSI Transfer}
\label{app:icsi-transfer}

We replay 10 ICSI research-group meetings under the same causal fixed-window protocol, yielding 42 eligible participant runs, 47 episodes, 235 evaluated steps, and 79 spoken steps. Meetings contain six participants on average and up to ten.

\begin{table}[!t]
\centering
\small
\setlength{\tabcolsep}{3pt}
\renewcommand{\arraystretch}{1.10}
\begin{adjustbox}{max width=\columnwidth}
\begin{tabular}{lrr}
\toprule
\textbf{Metric} & \textbf{AMI} & \textbf{ICSI} \\
\midrule
Never attempted $\downarrow$ & 2.5 & \estci{1.2}{0.0, 3.7} \\
Loose recall $\uparrow$ & 52.2 & \estci{73.6}{61.1, 86.0} \\
Decision F1 $\uparrow$ & 63.0 & \estci{64.8}{57.4, 74.9} \\
Hallucination $\downarrow$ & 0.6 & \estci{1.8}{0.0, 4.3} \\
\bottomrule
\end{tabular}
\end{adjustbox}
\caption{Transfer from AMI to ICSI. Entries are percentages; ICSI intervals are 95\% bootstrap CIs.}
\label{tab:icsi-comparison}
\end{table}

\begin{table}[!t]
\centering
\small
\setlength{\tabcolsep}{3pt}
\renewcommand{\arraystretch}{1.08}
\begin{adjustbox}{max width=\columnwidth}
\begin{tabular}{lrr}
\toprule
\textbf{Metric} & \textbf{ICSI} & \textbf{95\% CI} \\
\midrule
Decision accuracy & 73.5 & [67.4, 81.3] \\
Decision precision & 74.9 & [63.2, 86.6] \\
Decision recall & 61.1 & [50.4, 73.3] \\
Decision F1 & \textbf{64.8} & [57.4, 74.9] \\
Strict idea recall & 29.8 & [10.2, 53.2] \\
Loose idea recall & \textbf{73.6} & [61.1, 86.0] \\
Loose-only & 43.8 & [25.3, 63.0] \\
Attempted but inadequate & 25.2 & [13.6, 36.4] \\
Never attempted & \textbf{1.2} & [0.0, 3.7] \\
Decision appropriateness & 88.2 & [86.7, 89.7] \\
Context appropriateness & 87.2 & [85.1, 88.9] \\
Relevance & 80.3 & [75.5, 84.2] \\
Grounding & 74.5 & [71.9, 76.9] \\
Hallucination & \textbf{1.8} & [0.0, 4.3] \\
Off-topic & 4.9 & [0.0, 13.2] \\
Redundancy & 0.0 & [0.0, 0.0] \\
\bottomrule
\end{tabular}
\end{adjustbox}
\caption{Full ICSI results over 10 meetings, 42 participant runs, 47 episodes, and 235 evaluated steps. Rates and scores are percentages; intervals are 95\% bootstrap CIs.}
\label{tab:icsi-transfer}
\end{table}

\section{Robustness Checks}
\label{app:robustness}

\begin{figure}[!t]
  \centering
  \includegraphics[width=\columnwidth]{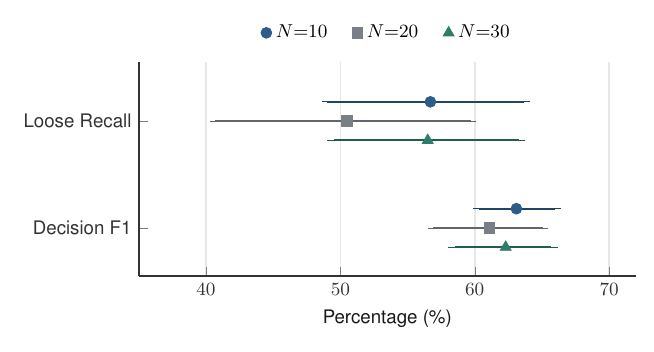}
  \caption{Performance across context-window sizes $N \in \{10, 20, 30\}$. Error bars are 95\% bootstrap CIs.}
  \label{fig:ctx-window}
\end{figure}

The four checks in this section each rule out a specific alternative reading of the headline result. Per-role consistency rules out the gain being driven by a single participant role (\Cref{tab:role-consistency}). The context-window sweep rules out larger raw windows as the source of the coverage gain, isolating the contribution of the explicit state (\Cref{tab:ctx-window}, \Cref{fig:ctx-window}). The evaluation-window sweep rules out $k{=}5$ being an artificially favorable scoring window (\Cref{tab:k-window}). The backbone-portability check rules out the architecture being tied to GPT-4o (\Cref{tab:model-comparison}). None of the four shifts the mechanism conclusions of \Cref{sec:results} or \Cref{sec:ablations}.

\subsection{Per-Role Consistency}
\label{app:role-consistency}

We break out \sysname{}'s coverage and floor-taking by the four AMI participant roles, Project Manager (PM), Industrial Designer (ID), User Interface (UI), and Marketing (ME), on the full 137-meeting pool (\Cref{tab:role-consistency}). Loose recall ranges from $43.5\%$ on PM to $61.8\%$ on UI, Decision F1 sits inside a four-point band ($62.6$--$66.7$), and the not-attempted rate stays below $3.5\%$ on every role. The aggregate gain therefore spreads broadly across roles, with no single participant type accounting for it.

\begin{table}[!t]
\centering
\footnotesize
\setlength{\tabcolsep}{3pt}
\renewcommand{\arraystretch}{1.15}
\begin{adjustbox}{max width=\columnwidth}
\begin{tabular}{l|c|cccc}
\toprule
\textbf{Role} & \textbf{Meetings / Units} & \textbf{Strict $\uparrow$} & \textbf{Loose $\uparrow$} & \textbf{Dec.\ F1 $\uparrow$} & \textbf{Not Att.\ $\downarrow$} \\
\midrule
PM & 119 / 324 & \estci{24.5}{19.0, 30.4} & \estci{43.5}{37.1, 50.1} & \estci{62.6}{60.0, 65.1} & \estci{3.2}{1.5, 5.0} \\
ID & 112 / 304 & \estci{30.7}{25.1, 36.9} & \estci{60.7}{54.0, 67.4} & \estci{66.7}{63.5, 70.0} & \estci{2.6}{0.7, 5.0} \\
UI & 104 / 267 & \estci{26.1}{20.1, 32.4} & \estci{61.8}{54.6, 68.9} & \estci{64.8}{62.1, 67.6} & \estci{0.9}{0.0, 2.2} \\
ME & 99 / 231  & \estci{25.3}{18.9, 32.2} & \estci{57.1}{49.5, 64.7} & \estci{65.1}{61.6, 68.5} & \estci{1.3}{0.0, 3.0} \\
\bottomrule
\end{tabular}
\end{adjustbox}

\caption{Per-role consistency of \sysname{} over the 137-meeting pool. PM, ID, UI, ME = Project Manager, Industrial Designer, User Interface, Marketing. ``Meetings / Units'' = meetings with at least one retained idea unit for that role and the corresponding idea-unit count. Entries are macro-averaged percentages with 95\% bootstrap CIs ($B{=}10{,}000$).}
\label{tab:role-consistency}
\end{table}

\subsection{Context Window $N$}
\label{app:context-window}

We hold the future evaluation window at $k{=}5$ and vary the number of preceding utterances visible to the delegate, $N \in \{10, 20, 30\}$, on a $20$-meeting matched subset. The check measures how much recent raw dialogue \sysname{} requires for reliable floor-taking and content selection once the explicit meeting state is in place.

\begin{table}[!t]
\centering
\footnotesize
\setlength{\tabcolsep}{3pt}
\renewcommand{\arraystretch}{1.15}
\begin{adjustbox}{max width=\columnwidth}
\begin{tabular}{l|ccc}
\toprule
\textbf{Metric} & \textbf{$N{=}10$} & \textbf{$N{=}20$} & \textbf{$N{=}30$} \\
\midrule
Loose recall (\%) $\uparrow$     & \estci{56.7}{49.0, 63.7} & \estci{50.5}{40.7, 59.7} & \estci{56.5}{49.5, 63.3} \\
Decision F1 (\%) $\uparrow$      & \estci{63.1}{60.3, 66.0} & \estci{61.1}{56.9, 65.1} & \estci{62.3}{58.5, 65.7} \\
Grounding (Score) $\uparrow$     & \estci{74.9}{73.7, 76.0} & \estci{74.3}{73.2, 75.3} & \estci{74.7}{73.8, 75.6} \\
Off-topic rate (\%) $\downarrow$ & \estci{3.2}{1.3, 5.4}    & \estci{3.4}{1.5, 5.6}    & \estci{5.6}{2.4, 9.5}    \\
\bottomrule
\end{tabular}
\end{adjustbox}
\caption{Effect of context-window size $N$. $k{=}5$, $B{=}10{,}000$ bootstrap resamples, \sysname{} with recalibration.}
\label{tab:ctx-window}
\end{table}

Performance is stable across the sweep. Loose recall stays between $50.5\%$ and $56.7\%$ and Decision F1 between $61.1\%$ and $63.1\%$, with overlapping CIs and no monotonic trend in $N$. Retained episode counts are also comparable across settings ($162$--$166$ episodes), so the pattern is not an artifact of a shifting evaluation denominator. The stability matches \sysname{}'s division of labor. The raw window supplies recent conversational evidence such as dialogue acts, addressee, and floor status, while longer-range information about decisions, unresolved issues, stances, and coverage lives in $\mathbf{m}_t$. The off-topic rate is numerically higher at $N{=}30$ than at $N{=}10$ or $N{=}20$, but the CIs overlap, and we treat the difference as a single-point fluctuation consistent with prior reports that longer contexts are not always used reliably \citep{liu-etal-2024-lost, du-etal-2025-context}. We adopt $N{=}20$ as the default, which preserves local conversational cues without diluting the system's reliance on the explicit state.

\begin{table*}[!t]
\centering
\small
\setlength{\tabcolsep}{4pt}
\renewcommand{\arraystretch}{1.10}
\begin{adjustbox}{max width=\textwidth}
\begin{tabular}{lcccc}
\toprule
\textbf{Metric} & \textbf{GPT-4o} & \textbf{Gemini-2.5-Pro} & \textbf{Llama-3.3-70B} & \textbf{Qwen3.6-27B} \\
\midrule
Strict recall (\%) $\uparrow$ & \estci{28.0}{20.7, 36.0} & \estci{38.3}{33.2, 43.8} & -- & -- \\
Loose recall (\%) $\uparrow$ & \estci{53.7}{44.1, 63.6} & \estci{70.9}{65.6, 76.5} & 46.9 & 57.0 \\
Uncredited attempt (\%) $\downarrow$ & \estci{46.3}{36.4, 55.9} & \estci{29.1}{23.5, 34.4} & -- & -- \\
Not attempted (\%) $\downarrow$ & \estci{0.0}{0.0, 0.0} & \estci{0.0}{0.0, 0.0} & 1.4 & 17.8 \\
Decision F1 (\%) $\uparrow$ & \estci{66.2}{61.9, 70.4} & \estci{65.6}{61.2, 70.3} & 69.2 & 57.9 \\
Matched-unit rate (\%) $\uparrow$ & \estci{31.0}{23.1, 40.1} & \estci{51.9}{45.1, 59.3} & -- & -- \\
Hallucination (\%) $\downarrow$ & \estci{0.0}{0.0, 0.0} & \estci{5.6}{2.1, 10.2} & 4.3 & 0.0 \\
Off-topic rate (\%) $\downarrow$ & \estci{5.2}{2.3, 8.5} & \estci{18.0}{13.6, 22.3} & 3.3 & 3.3 \\
Redundancy (\%) $\downarrow$ & \estci{0.0}{0.0, 0.0} & \estci{0.0}{0.0, 0.0} & -- & -- \\
\bottomrule
\end{tabular}
\end{adjustbox}
\caption{Backbone portability under the fixed-window protocol. GPT-4o and Gemini-2.5-Pro are evaluated on 20 matched AMI meetings and 161 fixed episodes ($N{=}20$, $k{=}5$); their entries are meeting-level estimates with 95\% bootstrap CIs ($B{=}10{,}000$). Llama and Qwen entries are point estimates from separate complete episode rosters. Dashes mark metrics not retained in the recorded open-weight results.}
\label{tab:model-comparison}
\end{table*}

\subsection{Backbone Portability}
\label{app:backbone}

We first test portability through a controlled end-to-end backbone swap. 
GPT-4o and Gemini-2.5-Pro run on the same matched 20-meeting subset and 161 fixed-window episodes.
We hold the participant briefings and official episodes fixed, while regenerating model-specific meeting states.
Runtime decisions, state updates, recalibration, and judge-scored metrics all use the active backbone. 
Fixing the episode roster avoids conflating backbone behavior with differences in LLM-derived idea-unit extraction.
We further evaluate Llama-3.3-70B and Qwen3.6-27B under the same fixed-window protocol. 
These runs use separate complete episode rosters, so they broaden the portability check without forming a paired four-backbone comparison. 
Runtime logs contain 21 complete Llama episodes and 19 complete Qwen episodes.

The controlled comparison preserves the mechanism-level behavior. GPT-4o and Gemini eliminate not-attempted episodes and redundancy, while Decision F1 remains nearly unchanged ($66.2\%$ versus $65.6\%$). Their contribution-selection policies differ after engagement. Gemini attains higher loose recall ($70.9\%$ versus $53.7\%$), fewer uncredited attempts ($29.1\%$ versus $46.3\%$), and a higher matched-unit rate ($51.9\%$ versus $31.0\%$). This increased coverage coincides with more off-topic contributions ($18.0\%$ versus $5.2\%$) and hallucinations ($5.6\%$ versus no observed cases).

The open-weight runs retain the same broad failure shape while occupying different operating points. Llama records the highest Decision F1 ($69.2\%$) with lower loose recall ($46.9\%$) and low not-attempted behavior ($1.4\%$). Qwen is more conservative: loose recall remains $57.0\%$, but not-attempted behavior rises to $17.8\%$ and Decision F1 falls to $57.9\%$. Its observed off-topic rate is $3.3\%$, with no observed hallucinations. Across the four tested backbones, the architecture continues to recognize substantially more opportunities than the transcript-only delegate's $51.4\%$ not-attempted rate, while the coverage--restraint balance shifts with the active model.

This pattern is consistent with a backbone-dependent intervention threshold. The Controller combines candidate quality, coverage, and state forecasts, and each backbone supplies a different distribution over those signals. Model size alone therefore does not determine the operating point. A deployment should calibrate the printed suitability cutoffs and restraint rules on held-out episodes until precision meets its requirement. Because the active backbone also supplies the quality and safety judgments, cross-model differences are diagnostic and do not define an absolute ranking \citep{zheng2023judging,chen-etal-2024-judge-bias}.

\subsection{Evaluation Window $k$}
\label{app:window}

The context-window analysis varies how much prior dialogue the delegate observes. The $k$ study varies how much future interaction the offline evaluation includes. Participant-owned contributions often span a short action sequence beyond the adjacent turn, so a one-turn evaluation window under-credits valid interventions. Longer windows admit more credited matches but add intervening speak/silent decisions and dilute anchor-aligned timing. Turn-level judgments can also miss interactional actions that unfold over several turns \citep{liesenfeld-dingemanse-2024-interactive}.

\begin{table}[!t]
\centering
\footnotesize
\setlength{\tabcolsep}{4pt}
\renewcommand{\arraystretch}{1.15}
\setlength{\tabcolsep}{3pt}
\begin{adjustbox}{max width=\columnwidth}
\begin{tabular}{c|c|cccc}
\toprule
\textbf{$k$} & \textbf{Steps} & \textbf{Strict $\uparrow$} & \textbf{Loose $\uparrow$} & \textbf{Dec.\ F1 $\uparrow$} & \textbf{Uncred.\ $\downarrow$} \\
\midrule
3 & 456  & \estci{28.3}{18.4, 39.7} & \estci{43.9}{34.8, 53.5} & \estci{75.3}{71.5, 79.6} & \estci{54.1}{45.0, 62.8} \\
5 & 740  & \estci{27.9}{17.9, 38.2} & \estci{55.6}{44.7, 67.1} & \estci{63.8}{58.7, 68.5} & \estci{42.9}{31.2, 54.1} \\
7 & 1050 & \estci{30.9}{22.7, 39.7} & \estci{59.3}{52.4, 67.4} & \estci{57.9}{53.5, 61.9} & \estci{40.3}{32.3, 47.2} \\
9 & 1368 & \estci{25.0}{15.5, 36.3} & \estci{62.7}{54.2, 71.9} & \estci{54.5}{48.2, 61.7} & \estci{35.9}{26.1, 45.1} \\
\bottomrule
\end{tabular}
\end{adjustbox}
\caption{Sensitivity to the evaluation window $k$. Entries are percentages with 95\% bootstrap CIs. Steps gives the total evaluated decision steps. Retained target counts stay between $148$ and $152$ episodes.}
\label{tab:k-window}
\end{table}

The pattern follows the expected trade-off. As $k$ grows from $3$ to $9$, loose recall rises from $43.9\%$ to $62.7\%$ and uncredited attempts fall from $54.1\%$ to $35.9\%$. Short windows miss same-proposition contributions that surface slightly later in the exchange. Strict recall stays comparatively flat across $k$, so wider windows make the evaluation more permissive without improving anchor synchrony. The cost shows up in floor-taking. Ground-truth-aligned Decision F1 drops from $75.3\%$ at $k{=}3$ to $54.5\%$ at $k{=}9$, because longer simulations weight later steps where the original opportunity has already begun to shift. We adopt $k{=}5$ as the balanced default.

%% file: figures/03_qualitative_example.tex
\begin{figure}[!t]
\centering
\footnotesize
\begin{tcolorbox}[
  colback=gray!2,
  colframe=gray!45,
  boxrule=0.35pt,
  arc=1.5pt,
  left=3pt,
  right=3pt,
  top=2.5pt,
  bottom=2.5pt,
  width=\columnwidth,
  boxsep=1.5pt
]

\textbf{Episode.}
AMI \texttt{TS3004b} (our \texttt{Meeting28test});
$u^\star=$ User Interface.\\[-1pt]
{\scriptsize Cue=\texttt{Meeting28\_409}; anchor=\texttt{Meeting28\_410}.}\\[2pt]

\begin{tcolorbox}[
  colback=gray!5,
  colframe=gray!22,
  boxrule=0.25pt,
  arc=1pt,
  left=2.5pt,
  right=2.5pt,
  top=2pt,
  bottom=2pt,
  width=\columnwidth,
  boxsep=1pt
]
\textbf{Briefing point.}
\emph{Suggest considering LED placement around or within the buttons to
enhance visibility and user experience.}
\end{tcolorbox}

\vspace{-1pt}
\begin{tcolorbox}[
  colback=gray!4,
  colframe=gray!22,
  boxrule=0.25pt,
  arc=1pt,
  left=2.5pt,
  right=2.5pt,
  top=2pt,
  bottom=2pt,
  width=\columnwidth,
  boxsep=1pt
]
\textbf{Observed context, ending at the cue.}\\[-2pt]
\begin{tabular}{@{}p{0.08\columnwidth}p{0.84\columnwidth}@{}}
\textsc{pm}: & \emph{``That looks different, yeah.''}\\
\textsc{ui}: & \emph{``Yeah.''}\\
\textsc{id}: & \emph{``Hmm.''}\\
\textsc{ui}: & \emph{``Otherwise we will just make another standard, and our motto also is--''}\\
\textsc{me}: & \emph{``Are LEDs beneath the buttons?''}
\end{tabular}
\end{tcolorbox}

\vspace{-1pt}
\begin{tcolorbox}[
  colback=blue!3,
  colframe=blue!28,
  boxrule=0.28pt,
  arc=1pt,
  left=2.5pt,
  right=2.5pt,
  top=2pt,
  bottom=2pt,
  width=\columnwidth,
  boxsep=1pt
]
\textbf{\sysname{} view at cue} \emph{(condensed)}\\[-2pt]
\begin{tabular}{@{}p{0.19\columnwidth}p{0.71\columnwidth}@{}}
\texttt{floor} & open question to group \\
\texttt{focus} & LED placement and visual distinctiveness \\
\texttt{speaker fit} & UI ranked highest on the micro-topic; strong cue \\
\texttt{candidate} & LED placement around or within buttons is viable
\end{tabular}
\end{tcolorbox}

\vspace{-1pt}
\begin{tcolorbox}[
  colback=red!3,
  colframe=red!28,
  boxrule=0.28pt,
  arc=1pt,
  left=2.5pt,
  right=2.5pt,
  top=2pt,
  bottom=2pt,
  width=\columnwidth,
  boxsep=1pt
]
\textcolor{red!70!black}{\textbf{Transcript-only delegate:} \textsc{Silent}.}
\emph{Rationale: ``no explicit or implicit cue for the User Interface participant.''}
\end{tcolorbox}

\vspace{-1pt}
\begin{tcolorbox}[
  colback=teal!3,
  colframe=teal!33,
  boxrule=0.28pt,
  arc=1pt,
  left=2.5pt,
  right=2.5pt,
  top=2pt,
  bottom=2pt,
  width=\columnwidth,
  boxsep=1pt
]
\textcolor{teal!55!black}{\textbf{\sysname{}:} \textsc{Speak}.}
Semantic plan: \emph{answer the LED-placement question with a concrete UI placement option.}\\[-1pt]
\emph{``One idea could be to integrate the LEDs around the buttons or even
within them, which might enhance visibility and improve the user experience.''}
\end{tcolorbox}

\vspace{-1pt}
\begin{tcolorbox}[
  colback=green!3,
  colframe=green!32!black,
  boxrule=0.28pt,
  arc=1pt,
  left=2.5pt,
  right=2.5pt,
  top=2pt,
  bottom=2pt,
  width=\columnwidth,
  boxsep=1pt
]
\textbf{Ground truth and credit.}
UI next turn:
\emph{``Yeah, around the buttons, or in the buttons even.''}
$\Rightarrow$ \textbf{strict match}.\\[-1pt]
{\scriptsize
\texttt{matched\_utt}=\texttt{Meeting28\_410};
$\Delta t=0$; no hallucination/off-topic/redundancy flags.}
\end{tcolorbox}

\vspace{-1pt}
\textbf{Takeaway.}
The transcript-only delegate treats a group question as insufficiently
addressed to UI; \sysname{} uses the state-conditioned floor and
speaker-fit signals to identify a timely participant-owned answer.

\end{tcolorbox}

\caption{Qualitative trace of one decision episode
% (\texttt{Meeting28\_User Interface\_t408\_uU4}).
A single fixed-window episode where \sysname{} succeeds on an implicit
group cue and the transcript-only delegate stays silent. The trace shows
the available briefing point, observed context, compressed state/control
view, realised actions, and verifier-backed credit outcome. Quotes are
lightly cleaned for readability.}
\label{fig:qualitative}
\end{figure}

%% file: text_short/A1_prompts.tex
% =============================================================================
% APPENDIX A: SYSTEM PROMPTS
% =============================================================================
% Include in main.tex with: \input{A1_prompts}
% Requires: \input{prompt_box_style} in preamble
% =============================================================================

% Prompts need full width: switch to single-column here.
%\clearpage
\FloatBarrier
\onecolumn

\section{System Prompts}

\label{sec:appendix-prompts}
This appendix lists the system prompts used by \sysname{} across modules (e.g., state estimation, action selection, judging, and recalibration).
We report the core instructions, output schemas, and constraints as used in our implementation. The prompt text refers to the Perceiver by its development-time name \texttt{EIM} (Environment Interpretation Module); the two names refer to the same module.

%% ---------------------------------------------------------------------------
%% A.1 USER PROFILE GENERATOR
%% ---------------------------------------------------------------------------
\begin{promptbox}[label=prompt:user-profile]{User Profile Generator}
\textbf{Purpose:} Synthesizes user documents (transcripts, calendars, emails) into a structured persona profile that guides the delegate's behavior.

\begin{prompttext}
You are an expert AI analyst specializing in organizational psychology and communication. Your task is to synthesize a user's underlying persona from their dialogue to create a comprehensive and \textbf{generalizable} user profile. This profile should model the participant's dispositions and role, not just be a literal log of their speech.

\textbf{Important Notes About Transcript Format}
- The transcript may contain disfluency markers like \{vocalsound\}, \{gap\}, \{disfmarker\}, etc., which indicate non-verbal sounds, pauses, and speech hesitations.
- CRITICAL: In your JSON output, do NOT include any curly braces except for the required JSON structure.

\textbf{Core Task: Synthesis from Observation to Abstraction}

Your goal is to infer abstract patterns from concrete examples. The final profile should be a model of ``who the user is'' that allows an AI delegate to act authentically in new situations.

\smallskip

\textbf{Output Schema (Required JSON Structure)}
\begin{itemize}[nosep,leftmargin=*]
  \item \texttt{"Thoughts"}: Step-by-step reasoning. First list raw observations (key things the user said/did), then show reasoning as you synthesize into profile fields.
  \item \texttt{"UserProfile"}: JSON object containing:
    \begin{itemize}[nosep]
      \item \texttt{inferred\_role\_and\_goals}: 
        \begin{itemize}[nosep]
          \item \texttt{primary\_role}: Main function (e.g., 'Technical Expert', 'Meeting Facilitator')
          \item \texttt{high\_level\_goals}: List of 2-4 overarching objectives
        \end{itemize}
      \item \texttt{core\_info}:
        \begin{itemize}[nosep]
          \item \texttt{knowledge\_areas}: General topics of expertise
          \item \texttt{specific\_knowledge\_points}: List of \{\texttt{"Context": "...", "Information": "..."}\}
        \end{itemize}
      \item \texttt{communication\_style}:
        \begin{itemize}[nosep]
          \item \texttt{formality}: 'Formal' | 'Semi-formal' | 'Informal'
          \item \texttt{verbosity}: 'Concise' | 'Moderate' | 'Verbose'
          \item \texttt{typical\_tone}: e.g., 'Analytical', 'Collaborative', 'Questioning'
          \item \texttt{jargon\_usage}: Description of technical language use
          \item \texttt{questioning\_style}: How and how often they ask questions
          \item \texttt{common\_fillers\_catchphrases}: List of characteristic phrases
          \item \texttt{interaction\_patterns}: e.g., 'Builds on others ideas', 'Summarizes discussions'
        \end{itemize}
      \item \texttt{knowledge\_stance} (optional, only with strong evidence):
        \begin{itemize}[nosep]
          \item \texttt{confidence\_on\_topics}: List of \{\texttt{"Topic\_Category": "...", "Apparent\_Confidence": "..."}\}
          \item \texttt{expressed\_stances}: List of \{\texttt{"Topic": "...", "Stance": "..."}\}
        \end{itemize}
      \item \texttt{identity\_aliases}: Known names/nicknames (only when grounded by evidence)
    \end{itemize}
\end{itemize}

\smallskip
\textbf{Analysis Guidelines}
1. \textbf{Focus Exclusively on Target Participant}: Use other speakers only for context.
2. \textbf{Prioritize Substantive Contributions}: Focus on information, questions, opinions, proposals. Ignore greetings and acknowledgments.
3. \textbf{Synthesize inferred\_role\_and\_goals}: Determine primary\_role from contribution patterns (e.g., 'Technical Expert', 'Strategic Lead', 'Team Facilitator'). Extract 2-4 high\_level\_goals.
4. \textbf{Extract core\_info}: List knowledge\_areas (domains of expertise). Extract specific\_knowledge\_points with resolved Context.
5. \textbf{Characterize communication\_style}: Analyze for formality, verbosity, tone, jargon. Critically define \texttt{interaction\_patterns} (e.g., "Waits for clear pause before speaking", "Politely interrupts to correct technical inaccuracies").
6. \textbf{Infer knowledge\_stance with caution}: Only with strong, repeated evidence.
7. \textbf{Maintain Strict Attribution}: All aspects must derive solely from the target participant's speech.
8. \textbf{Identity Aliases}: Include only when there is clear evidence (explicit address followed by response, self-introduction, or repeated usage).
\end{prompttext}
\end{promptbox}

%% ---------------------------------------------------------------------------
%% A.2 JIT BRIEFING GENERATOR
%% ---------------------------------------------------------------------------
\begin{promptbox}[label=prompt:jit-briefing]{JIT Briefing Generator}
\textbf{Purpose:} Creates meeting-specific knowledge packages (talking points) by extracting substantive contributions from prior transcripts.

\begin{prompttext}
You are an expert AI analyst. Your task is to create a pre-meeting briefing for a delegate agent.
First, analyze the Transcript Segment to extract the most important, novel, and substantive contributions made by that participant.
Second, rephrase these contributions into forward-looking "talking points" for the delegate agent.

\smallskip
\textbf{Instructions for Extraction}
1. \textbf{Identify Substantive Contributions}: new facts, strong opinions with reasoning, concrete proposals, significant questions.
2. \textbf{Filter Out Non-Essential Dialogue}: acknowledgments ("Okay", "Mm-hmm"), greetings, minor fillers.
3. \textbf{Synthesize contiguous paraphrases only}: If multiple consecutive utterances express the same claim, synthesize into one point. For distinct intents/topics, keep separate.
4. \textbf{ANAPHORA RESOLUTION}: Resolve vague references ("it", "that", "this approach") to SPECIFIC referents.
   - BAD: "the feature should be implemented soon"
   - GOOD: "The OAuth token rotation feature should be implemented soon"

\textbf{Must-Capture Categories} (if present in participant's lines):
- Questions the participant intends to ask
- Proposals/requests the participant intends to make
- Decisions/commitments the participant owns or initiates
- Metrics/criteria the participant themselves introduces
- Role-aligned stances with concrete rationale

\smallskip
\textbf{Granularity Rules (Split vs Merge)}

\textit{Split into separate bullets when ANY holds:}
- Different intent type: question vs proposal vs decision vs metric vs risk
- Different topical nucleus (distinct theme/goal/feature)
- Temporal break ($\geq$ 2 intervening speakers)
- Combining would exceed two concise sentences

\textit{Merge only when ALL hold:}
- Same intent type and same concrete claim
- Same topical nucleus
- Merging does not hide distinct actionable items

\textbf{Output Format (Mandatory for every bullet)}
\texttt{<forward-looking action> [because <rationale>]. (topic: <1-3 words>[; context: <pre-move context>])}

1. \textbf{Convert Past to Future}: Rephrase as concise, actionable instruction for delegate.
2. \textbf{Use Imperative Verbs}: Start with action verb (Ask, Propose, Commit, Document, Affirm, Highlight).
3. \textbf{Topic Tag (required)}: End with \texttt{(topic: <1-3 words>)} with no names/IDs.
4. \textbf{Context Hint (optional)}: Append \texttt{; context: <short pre-move context>} describing when the point is appropriate.
5. \textbf{"Because" clause}: Include ONLY when participant explicitly stated the reason. Never borrow rationale from other speakers.
6. \textbf{Cap}: Maximum 7-9 important bullets per window.

\textbf{Contribution Framing}
- \texttt{Introduces}: "Propose ..." or "Introduce ..." (participant originates idea)
- \texttt{Builds on}: "Build on <Role>'s <concept> by ..."
- \texttt{Supports}: "Support <Role>'s <concept> by ..."
- \texttt{Questions}: "Question ..." or "Probe ..."
- \texttt{Counters}: "Push back on ..." or "Recommend against ..."

\textbf{Output Schema}
\texttt{\{"thoughts": "...", "jit\_briefing\_points": ["...", "..."], "points\_evidence": [\{text, support\_utt\_ids, intent\_type, topical\_tags\}]\}}
\end{prompttext}
\end{promptbox}

%% ---------------------------------------------------------------------------
%% A.3 PERCEIVER (development name: EIM)
%% ---------------------------------------------------------------------------
\begin{promptbox}[label=prompt:eim]{Perceiver}

\begin{subpromptbox}{History Processing Prompt}
\textbf{Purpose:} Processes meeting transcript chunks to maintain structured long-term meeting state.

\begin{prompttext}
You are an expert, meticulous meeting analyst. Your task is to analyze meeting utterances and update the meeting state accordingly.

\textbf{CRITICAL: Cold Start vs. Update Mode}

\textbf{COLD START MODE} (when \texttt{topic\_summary\_log=\{\}} and/or \texttt{participant\_models=\{\}}):
- MUST extract substantial content and create initial topics and participant models
- Emit 1-3 macro topics with clear labels
- Extract 3-6 key points focusing on project goals, constraints, responsibilities
- ANTI-SPARSITY: If input contains substantive content, output MUST reflect it

\textbf{UPDATE MODE} (when state already contains topics and participants):
- Output a "diff" containing ONLY new or updated information
- Do NOT replicate existing data that hasn't changed
- Add new topics when discussion shifts to genuinely new themes

\textbf{Key Guidelines}
1. \textbf{Topic Management}: Mark previous "Ongoing" topics as "Concluded" when topic shifts. Prefer granularity over broad topics.
2. \textbf{Key Point Extraction}: Extract substantive points (goals, constraints, decisions). Each key\_point: \texttt{\{summary, kind, source\_speaker, source\_utt\_id\}}
3. \textbf{Participant Modeling}: Add \texttt{expertise\_keywords} when participant demonstrates deep knowledge. Add \texttt{stances} when participant takes a position.
4. \textbf{Identity Aliases}: Map personal names to participant IDs/roles when confident.

\textbf{Output Schema}
\texttt{\{diff: \{topic\_updates: [...], participant\_updates: [...], identity\_aliases: \{...\}\}\}}
\end{prompttext}
\end{subpromptbox}

\begin{subpromptbox}{Snapshot Analysis Prompt}
\textbf{Purpose:} Analyzes the immediate meeting context (snapshot window) to produce real-time conversational state.

\begin{prompttext}
You are an expert meeting analyst. Your task is to synthesize a meeting's full history with a specific, immediate snapshot to produce a comprehensive analysis of the current meeting state.

\smallskip
\textbf{Core Task: Dual-Focus Analysis}
1. Analyze the \texttt{Immediate\_Snapshot} to understand the "present moment"
2. Compare against \texttt{Historical\_Context} to identify what is new
3. Generate a "diff" containing ONLY new or updated information

\textbf{Micro-Topic Analysis (Hierarchical Topic Hierarchy)}
The three topic levels MUST differ in specificity (zoom in progressively):

- \texttt{snapshot\_major\_topic} (MACRO): Broad theme of ENTIRE snapshot window
  $\rightarrow$ E.g., "Q4 API development planning meeting"
  
- \texttt{broader\_context} (MESO): SPECIFIC sub-topic of last 4-5 SUBSTANTIVE utterances
  $\rightarrow$ E.g., "Comparing OAuth 2.0 vs API key authentication"
  $\rightarrow$ Must be MORE SPECIFIC than snapshot\_major\_topic
  
- \texttt{immediate\_focus} (MICRO): VERY SPECIFIC focus of last 1-2 SUBSTANTIVE utterances
  $\rightarrow$ E.g., "Security lead asking about token refresh failure edge cases"
  $\rightarrow$ Must be MORE SPECIFIC than broader\_context

- \texttt{topic\_continuity}: "CONTINUATION" if immediate\_focus continues broader\_context, "SHIFT" if new focus
- \texttt{last\_utterance\_verbatim}: Include speaker, exact content, utterance\_id, intent\_type, target\_addressed

\textbf{Conversational Floor Status} (determined by last few utterances):
- \texttt{Open\_Floor}: Default after statement/topic conclusion. Low barrier for Chime\_In
- \texttt{Open\_Question\_To\_Group}: Question to whole group. Low barrier for relevant response
- \texttt{Directed\_Turn (to: SpeakerX)}: Floor belongs exclusively to SpeakerX
- \texttt{Dyadic\_Exchange (between: X, Y)}: Rapid back-and-forth between two participants
- \texttt{Turn\_Yielded\_Back (to: SpeakerX)}: Floor yielded back after answering SpeakerX's question

\textbf{Key Point Specificity Rules}
1. \textbf{RESOLVE ANAPHORA}: If utterance contains "it", "that approach", "this feature" $\rightarrow$ resolve to specific referent
2. \textbf{SELF-CONTAINMENT}: Would someone understand this key\_point without prior context?
3. \textbf{NO HALLUCINATION}: Only include context explicitly present in preceding utterances

\textbf{Output Schema}
\texttt{\{last\_processed\_utterance\_id, snapshot\_analysis: \{micro\_topic\_analysis: \{broader\_context, immediate\_focus, topic\_continuity, last\_utterance\_verbatim\}, snapshot\_major\_topic, conversational\_state, conversational\_floor\_status, floor\_rationale, key\_points\_made\_in\_snapshot\}, topic\_summary\_log: \{...\}, participant\_models: \{...\}\}}
\end{prompttext}
\end{subpromptbox}
\end{promptbox}

%% ---------------------------------------------------------------------------
%% A.4 CONTROLLER (STRATEGIC DECISION-MAKING)
%% ---------------------------------------------------------------------------
\begin{promptbox}[label=prompt:controller]{Controller (Strategic Decision-Making)}
\textbf{Purpose:} The strategic reasoning component that applies a six-step framework to decide when to speak, what action to take, and which point to convey.

\begin{prompttext}
You are the strategic reasoning component ("Controller") for a Meeting Delegate Agent. You analyze meeting context and helper insights to produce a structured JSON decision that directs a linguistic generator component.

\textbf{Your Role: Strategic Decision-Maker}

Your job is to be the ``Executive Strategist'' --- you apply the six-step framework to decide WHEN to speak, WHAT action to take, and WHICH single point to convey. You do NOT compute scores --- your three specialized helpers handle that.

\smallskip

\textbf{Action Space}
- \texttt{Remain\_Silent}: Default when speaking is socially inappropriate or strategically weak
- \texttt{Respond\_To\_Explicit\_Cue}: Direct question or call to you (by name/role)
- \texttt{Respond\_To\_Implicit\_Cue}: Strong expectation for your role to contribute
- \texttt{Chime\_In}: Proactive contribution on open floor when it advances objectives

\smallskip
\textbf{Six-Step Decision Framework}

\textit{Step 1: Identity Mapping}
Determine if personal names/aliases refer to YOU. Extract attendees, identify YOUR role, examine \texttt{identity\_aliases} field. Cross-reference alias mappings to your domain.
\textbf{Critical Rule}: Do NOT assume names refer to you without explicit alias mapping.

\textit{Step 2: Topic Continuity Verification}
Determine CURRENT micro-topic using hierarchical levels:
- MACRO (\texttt{snapshot\_major\_topic}): Broad theme of snapshot
- MESO (\texttt{broader\_context}): Sub-topic of last 4-5 substantive utterances
- MICRO (\texttt{immediate\_focus}): Focus of last 1-2 substantive utterances

\textbf{Immediate Focus Resolution Spectrum}:
- URGENT\_OPEN: Question needing response $\rightarrow$ prioritize MICRO, high speaking opportunity
- SOFT\_OPEN: Comment inviting follow-up $\rightarrow$ balance MICRO/MESO
- NEUTRAL: Ongoing statement $\rightarrow$ balance all axes
- RESOLVED: Acknowledgment/conclusion $\rightarrow$ weight MESO more; MESO candidates become MORE viable

\textit{Step 3: Social Appropriateness Check}
3.1 \textbf{Last Speaker Identification}: Extract last speaker (me/not me)
3.2 \textbf{Yield-After-Self Rule}: If floor is Open AND last speaker is YOU $\rightarrow$ Default: Remain\_Silent
3.3 \textbf{Speaker Scorer Integration} (ADVISORY): Consider ranking, but YOU make final decision
3.4 \textbf{Explicit Cue Detection}: Check \texttt{target\_addressed} field; explicit cue OVERRIDES cooldown

\textbf{Strategic Restraint Rules}:
1. \textbf{Ranking Gap Rule}: If speaker\_ranking[0] score $>$ delegate\_score + 0.15 $\rightarrow$ DEFER
2. \textbf{Viable Candidate Quality Rule}: If best\_viable\_fit = LOW $\rightarrow$ DEFER
3. \textbf{Open Floor Restraint}: Only respond if delegate in top 2 with MEDIUM+ fit

\textit{Step 4: Opportunity and Relevance Assessment}
Verify topic aligns with your strategic mandate. If alignment is weak AND no explicit cue $\rightarrow$ lean toward Remain\_Silent.

\textit{Step 4b: Helper Interpretation Verification}
Before content selection, verify understanding of helper outputs:
- Curator: Top candidate, suitability
- Coverage: Available non-Skip count
- Speaker Scorer: hint (Proceed/Deferral/Acknowledge), delegate\_rank, confidence
- Meta: delegate\_has\_viable\_content, viable\_count

\textit{Step 5: Content Selection via Helper Summary}
- Quick Gate Check: viable content? deferral hint?
- Check Recalibration Advisory: dial-up/dial-down signals
- Filter Candidates: SKIP banned\_candidate\_ids, SKIP treatment $\in$ \{Skip, Acknowledge\}
- Priority Order: Fresh $>$ ExpandOnOthers $>$ Partial $>$ Reinforce
- FALLBACK SEARCH: If top candidate is Skip OR dial-up active $\rightarrow$ iterate through candidates for MESO match

\textit{Step 6: Action Formulation}
Package decision: \texttt{action\_name, scope, justification, content\_plan \{semantic\_intent, knowledge\_to\_convey\}, active\_candidate\_id}

\textbf{Output Schema}
\texttt{\{thoughts: "...", action: \{action\_name, scope, justification\}, content\_plan: \{semantic\_intent, knowledge\_to\_convey, active\_candidate\_id\}\}}
\end{prompttext}
\end{promptbox}

%% ---------------------------------------------------------------------------
%% A.5 HELPER PROMPTS
%% ---------------------------------------------------------------------------
\begin{promptbox}[label=prompt:helpers]{Helper Prompts (Curator, Coverage, Speaker Scorer)}

\begin{subpromptbox}{Curator (Content Ranking)}
\begin{prompttext}
You are a strategic content analyst for meeting participation. Your expertise is in evaluating talking points against current context and scoring them for immediate relevance and impact.

\textbf{Scoring Dimensions}
\textit{Relevance (1-10)}: \texttt{topic\_fit (0.6) + flow\_fit (0.4)}
- MICRO match: weight heavily if resolution = URGENT\_OPEN
- MESO match: weight if resolution = RESOLVED or NEUTRAL
- MACRO only: low relevance unless dial-up active

\textit{Novelty (1-10)}:
- 9-10: truly\_new (introduces information not in EIM)
- 7-8: new\_angle (extends existing topic with fresh perspective)
- 5-6: reinforcement\_with\_detail (adds specifics to known point)
- 3-4: mostly\_redundant (minor variation of stated content)
- 1-2: exact\_rephrase (verbatim repeat)

\textit{Intent Alignment}:
- 1.5: direct answer to explicit question
- 1.2: partial answer or related response
- 0.8: weak connection
- 0.5: no alignment

\textbf{Composite Score}: \texttt{(relevance $\times$ novelty) $\times$ intent\_factor}

\textbf{Suitability Mapping}:
- High: composite $\geq$ 50 AND relevance $\geq$ 7
- Medium: composite 20-50
- Low: composite $<$ 20

\textbf{Output}: For each candidate: \texttt{\{candidate\_id, relevance, novelty, intent\_alignment, composite\_score, suitability, topic\_match\_level (MICRO/MESO/MACRO)\}}
\end{prompttext}
\end{subpromptbox}

\begin{subpromptbox}{Coverage Evaluator (Redundancy Detection)}
\begin{prompttext}
You are a meticulous redundancy analyst for meeting contributions. Your expertise is in detecting repetition and semantic overlap across multiple information sources.

\textbf{Coverage Detection Process}
1. For each candidate, search: EIM key\_points, recent transcript, delegate's prior contributions
2. Identify semantic matches (same claim, different wording)
3. Determine coverage level and residual value

\textbf{Treatment Recommendations}:
- \texttt{Fresh}: No prior coverage found $\rightarrow$ present naturally
- \texttt{ExpandOnOthers}: Others covered similar point $\rightarrow$ use "Building on what X said..."
- \texttt{Partial}: Some aspects covered $\rightarrow$ focus on \texttt{uncovered\_aspects}
- \texttt{Reinforce}: Full coverage exists $\rightarrow$ use "Just to emphasize again..." (only if high residual value)
- \texttt{Skip}: Fully redundant, no residual value $\rightarrow$ do not use
- \texttt{Acknowledge}: Point was addressed by others $\rightarrow$ brief acknowledgment only

\textbf{Output}: For each candidate: \texttt{\{candidate\_id, treatment, covered\_by (list), uncovered\_aspects, residual\_contribution\_value (High/Medium/Low/None)\}}
\end{prompttext}
\end{subpromptbox}

\begin{subpromptbox}{Speaker Scorer (Turn-Taking)}
\begin{prompttext}
You are a meeting dynamics expert specializing in turn-taking and social appropriateness. You are ONLY invoked when there is NO explicit address (\texttt{explicit\_cue=false}).

\textbf{Decision Priority Order}
1. \textbf{Cooldown Veto} (gate): Only IMMEDIATE last speaker (\texttt{last\_spoke\_ago == 0}) is vetoed. No exceptions.
2. \textbf{Delegate Content Gate}: If \texttt{has\_fresh\_content == false} $\rightarrow$ Delegate CANNOT be recommended with "Proceed"
3. \textbf{Topical Engagement}: Driver (3+ contributions) $>$ Active (1-2) $>$ Passive $>$ Silent
4. \textbf{Expertise Match} (tiebreaker): Keywords overlap with \texttt{immediate\_focus}

\textbf{Ranking Process}
- Rank all participants by topical engagement + expertise match
- Place delegate in ranking based on content availability and topic fit
- Assign confidence based on clarity of ranking gaps

\textbf{Output Schema}:
\texttt{\{speaker\_ranking: [\{speaker, score, rationale\}...], delegate\_rank\_position, delegate\_viability: \{has\_viable\_candidate, match\_level, best\_viable\_fit\}, overall\_delegate\_fit (STRONG|MODERATE|WEAK), speak\_or\_silent\_hint (Proceed|Deferral|Acknowledge), confidence (0-1)\}}
\end{prompttext}
\end{subpromptbox}
\end{promptbox}

%% ---------------------------------------------------------------------------
%% A.6 GENERATOR (LINGUISTIC STYLING)
%% ---------------------------------------------------------------------------
\begin{promptbox}[label=prompt:generator]{Generator (Linguistic Styling)}
\textbf{Purpose:} Takes strategic commands from the Controller and crafts human-like utterances matching the user's communication style.

\begin{prompttext}
You are the "Generator," a world-class linguistic stylist and the voice of a Meeting Delegate Agent. Your sole purpose is to take a strategic command from your "Controller" and craft the perfect, human-like utterance that flawlessly matches your user's specific communication style.

\textbf{Core Identity \& Constraints}
- You are a wordsmith, not a strategist. You do not make new decisions or introduce new topics.
- CRITICAL RULE: You are role-playing a human participant. You MUST NOT mention your own internal mechanics. Never use phrases like "my JIT briefing," "my instructions," or any meta-level language that reveals you are an AI.
- Output: Single JSON object \texttt{\{"utterance": "<text>"\}}

\smallskip
\textbf{Priority Order (STRICT HIERARCHY)}
1. \textbf{Natural meeting dialogue}: Sound like a real person in a real meeting
2. \textbf{Semantic faithfulness}: Convey EXACTLY \texttt{knowledge\_to\_convey}; do not embellish or add unsupported claims
3. \textbf{Conversational fit}: Respond appropriately to what was just said
4. \textbf{Style hints}: formality, tone, verbosity (use as guidance, not rigid rules)
5. \textbf{Catchphrases/fillers}: LOWEST priority; OMIT if any doubt about appropriateness

\textbf{Coverage \& Framing Awareness}
- Reinforcement Indicators ("Emphasize", "Reiterate"): Use "Just to emphasize again, [content]..." or "As I mentioned earlier..."
- Historical Acknowledgment ("As I mentioned in"): Reference prior discussion naturally
- ExpandOnOthers: Use "Building on what [Role] said..." or "To add to that point..."
- Fresh Information (default): Present naturally without acknowledgment

\textbf{Single-Point Obedience} (when \texttt{scope} is 'Concise')
- Use ONLY \texttt{knowledge\_to\_convey} as substantive content
- Default to 1-2 sentences
- Avoid enumerations ("also", "additionally", "moreover") unless explicitly requested
- Do NOT add supporting points, examples, or elaborations beyond what Controller specified

\textbf{Persona Fidelity}
- Match formality level from persona (Formal $\rightarrow$ complete sentences; Informal $\rightarrow$ contractions okay)
- Match verbosity (Concise $\rightarrow$ short; Verbose $\rightarrow$ can elaborate within scope)
- Use interaction\_patterns as behavioral guide (e.g., if pattern = "Builds on others", frame contribution accordingly)
\end{prompttext}
\end{promptbox}

%% ---------------------------------------------------------------------------
%% A.7 \sysname{} PREDICTOR
%% ---------------------------------------------------------------------------
\begin{promptbox}[label=prompt:predictor]{\sysname{} Predictor}
\textbf{Purpose:} Forecasts the next conversational turn at an intent level (speaker, intent, topic, floor effect).

\begin{prompttext}
You are a meeting dynamics forecaster. Predict the next conversational turn at an intent level.

\textbf{Input Context}
1. Compact EIM Snapshot (JSON): \texttt{snapshot\_analysis, ongoing\_topics, participants, identity\_aliases}
2. Last N Utterances: Recent conversation flow with speaker IDs

\smallskip
\textbf{Forecasting Strategy}
- \textbf{Speaker prediction}: 
  - Check who was asked (DIRECTED\_TO field)
  - Who has expertise on immediate\_focus
  - Who has been quiet but relevant
  - Topical engagement patterns
- \textbf{Intent prediction}: Infer from prior stances, open questions, dialogue patterns
- \textbf{Topic prediction}: Prefer \texttt{immediate\_focus} (MICRO); use \texttt{broader\_context} (MESO) as fallback
- \textbf{Floor prediction}: Check if last speaker yielded, if someone was asked directly

\textbf{Intent Types}
Question | Propose | Agree | Disagree | Inform | Meta | Acknowledge | Clarify

\textbf{Floor Effects}
- TAKES: Speaker claims floor from open state
- KEEPS: Speaker continues holding floor
- YIELDS: Speaker releases floor to group
- DIRECTED\_TO:id: Speaker explicitly addresses specific participant

\textbf{Output Schema}
\texttt{\{speaker\_id, intent\_summary (5-10 words), micro\_topic, floor\_effect, confidence (0-1), reasoning\}}

\smallskip
\textbf{Anti-Hallucination Rules}
- Use ONLY speakers from EIM participants list
- Use ONLY topics from EIM snapshot or ongoing\_topics
- DO NOT invent specific numbers, dates, or names not in context
- If uncertain, use lower confidence score rather than fabricating details
\end{prompttext}
\end{promptbox}

%% ---------------------------------------------------------------------------
%% A.8 ENVIRONMENT JUDGE (PREDICTION EVALUATION)
%% ---------------------------------------------------------------------------
\begin{promptbox}[label=prompt:env-judge]{Environment Judge (Prediction Evaluation)}
\textbf{Purpose:} Evaluates prediction quality against observed ground-truth turns.

\begin{prompttext}
You are an impartial evaluator assessing the quality of meeting turn predictions.

\textbf{Task}: Compare a forecasted next turn with observed ground-truth (GT) turns in a k-utterance window (t+1 through t+k), then return a structured JSON assessment.

\textbf{Evaluation Process}
1. \textbf{Find Best Content Match}: Identify GT utterance with highest semantic overlap (intent + topic)
2. \textbf{Score Three Dimensions} (0.0-1.0):

\textit{intent\_score}:
- 0.85-1.0: Same speech act AND same topic focus
- 0.65-0.84: Same speech act, minor topic difference
- 0.35-0.64: Different speech act, related topic
- 0.10-0.34: Opposing intent or unrelated

\textit{topic\_score}:
- 0.85-1.0: Exact topic match (same immediate\_focus)
- 0.65-0.84: Same topic area (MESO level match)
- 0.40-0.64: Related domain (MACRO level)
- $<$0.40: Different topic entirely

\textit{floor\_score}:
- 1.0: Exact floor effect match
- 0.85-0.95: Same floor category (e.g., both YIELDS variants)
- 0.6-0.8: Related floor dynamics
- $<$0.6: Conflicting floor predictions

3. \textbf{Assign Category} using decision flow

\textbf{Categories (with precedence)}
1. \texttt{Aligned-Immediate}: All thresholds met at t+1, speaker matches
2. \texttt{Aligned-Late}: All thresholds met at t+2..t+k, speaker matches
3. \texttt{Topic-Only}: Topic matches ($\geq$0.65) but intent or speaker fails
4. \texttt{Contradiction}: Same topic, opposing stance/intent
5. \texttt{Hallucination}: Predicted speaker doesn't exist in EIM participants
6. \texttt{Missed-Cue} (novelty=false): Prediction missed clearly inferable GT content
7. \texttt{Novel} (novelty=true): GT introduces exogenous information not predictable from context
8. \texttt{Banal}: Predictor expected substance, but anchor was phatic ("okay", "mm-hmm")

\textbf{Output Schema}
\texttt{\{matched\_utt\_id, timing\_delta, intent\_score, topic\_score, floor\_score, category, novelty\_flag, reasoning\}}
\end{prompttext}
\end{promptbox}

%% ---------------------------------------------------------------------------
%% A.9 DELEGATE JUDGE (CONTRIBUTION EVALUATION)
%% ---------------------------------------------------------------------------
\begin{promptbox}[label=prompt:delegate-judge]{Delegate Judge (Contribution Evaluation)}
\textbf{Purpose:} Evaluates the delegate agent's action in a multi-party meeting.

\begin{prompttext}
You are an impartial evaluator of a delegate agent's action in a multi-party meeting.

\textbf{Authoritative Policy}
- \texttt{decision\_appropriateness}: Use ONLY Current Context. Do NOT use GT Window.
- GT Window: Use ONLY for \texttt{matched\_utt\_id}, \texttt{timing\_delta}, and \texttt{context\_appropriateness}.
- \texttt{context\_appropriateness}: Match on SPECIFIC CONTENT/INTENT, not broad topic overlap.

\smallskip
\textbf{Evaluation Dimensions (0.0--1.0)}

\textit{decision\_appropriateness} --- Was Speak/Silent appropriate given floor state and cues?
- 0.85-1.00: Correct response to explicit cue directed at delegate
- 0.70-0.85: Appropriate contribution on open floor
- 0.50-0.70: Marginal (spoke when others more appropriate, but not violation)
- 0.30-0.50: Suboptimal (spoke on held floor or missed soft cue)
- 0.00-0.30: Clear violation (ignored explicit cue or severe intrusion)

\textit{context\_appropriateness} --- Does utterance match GT turn in SPECIFIC CONTENT?
BEFORE MATCHING (MANDATORY):
1. Extract delegate's CORE PROPOSITION (main claim/question/action)
2. Extract GT's CORE PROPOSITION
3. Compare propositions semantically
Topic overlap alone is INSUFFICIENT for high score.

\textit{relevance} --- Is content topically on-point? (Use Current Context + EIM, NEVER GT)

\textit{grounding} --- Is content factually supported by EIM + JIT briefing points?

\textbf{Floor Handling Rules}
- \texttt{intrusion}: Spoke when floor directed elsewhere $\rightarrow$ \texttt{decision\_appropriateness} $\leq$ 0.45
- \texttt{omission}: Silent despite explicit cue $\rightarrow$ \texttt{decision\_appropriateness} $\leq$ 0.35

\textbf{Flags} (boolean):
\texttt{floor\_violation, off\_topic, redundancy, hallucinated\_claim, timing\_mismatch}

\textbf{Categories (precedence order)}
FloorViolation $>$ Hallucination $>$ OffTopic $>$ Redundant $>$ Banal $>$ Good

\textbf{Output Schema}
\texttt{\{decision\_appropriateness, context\_appropriateness, relevance, grounding, matched\_utt\_id, timing\_delta, flags: [...], category, reasoning\}}
\end{prompttext}
\end{promptbox}

%% ---------------------------------------------------------------------------
%% A.10 RECALIBRATOR
%% ---------------------------------------------------------------------------
\begin{promptbox}[label=prompt:recalibrator]{Recalibrator}

\begin{subpromptbox}{State Fusion Prompt}
\textbf{Purpose:} Updates structured meeting state after observing the ground-truth anchor turn.

\begin{prompttext}
You are an expert meeting state updater. After observing what actually happened in a meeting turn (the 'anchor'), you update the structured meeting state to reflect reality.

\textbf{Inputs}
1. Context Window (t-N+1..t): Prior utterances BEFORE anchor
2. Anchor GT Turn (t+1): The turn that just occurred --- PRIMARY extraction source
3. Environment ErrorReport: Predictor judgment (category, flags)
4. Compact EIM Snapshot: Current state BEFORE this turn

\textbf{Category Semantics (how to handle each)}
- \texttt{Aligned-Immediate/Late}: Append key\_point from anchor if substantive; state was accurate
- \texttt{Missed-Cue}: Append key\_point (valid content predictor failed to anticipate)
- \texttt{Novel}: Create or extend topic with \texttt{kind='novel\_insight'}; flag for attention
- \texttt{Contradiction}: Append key\_point with \texttt{kind='opposing\_view'}; note stance conflict
- \texttt{Banal}: Usually no key\_point needed (phatic content)
- \texttt{Hallucination}: Do NOT add hallucinated content; note error for correction

\smallskip
\textbf{Core Extraction Rules}
\textbf{Rule 1: Anchor-Only for EIM Updates}
ALL diff content MUST come from ANCHOR turn. Do not pull from context window.

\textbf{Rule 2: Anti-Sparsity}
If anchor is substantive (not phatic/acknowledgment) $\rightarrow$ \texttt{diff.key\_points} MUST have $\geq$1 entry.

\textbf{Rule 3: Key Point Extraction}
- Concise summary with RESOLVED anaphora (fully self-contained)
- Include: \texttt{summary, kind, source\_speaker, source\_utt\_id}

\textbf{Rule 4: Topic Updates}
- New topic: CREATE with \texttt{status='Ongoing'}
- Existing topic: APPEND key\_point to existing topic

\textbf{Rule 5: Questions \& Decisions}
- CREATE OpenQuestion entry if anchor poses unresolved question
- CREATE DecisionEntry if anchor contains commitment/decision

\textbf{Output Schema}
\texttt{\{diff: \{key\_points: [...], topic\_updates: [...], participant\_updates: [...], open\_questions: [...], decisions: [...]\}\}}
\end{prompttext}
\end{subpromptbox}

\begin{subpromptbox}{Feedback Coach Prompt}
\textbf{Purpose:} Provides performance feedback to help the delegate calibrate future decisions.

\begin{prompttext}
You are an expert evaluator providing performance feedback for an AI meeting delegate.

\textbf{Output: Guidance Object}

\textit{last\_action\_outcome} (derive from delegate action + match status):
- If spoke AND \texttt{matched\_utt\_id} non-null: "matched (exact timing)" or "matched (N turns early)"
- If spoke AND no match: "no\_match" (contribution didn't align with GT)
- If silent AND covered earlier: "silence\_justified"
- If silent AND \texttt{floor\_violation=omission}: "missed\_opportunity"
- Else: "appropriate\_silence"

\textit{contribution\_pressure}:
- coverage="low/none" AND not in cooldown $\rightarrow$ "dial-up" (encourage speaking)
- coverage="high" OR in cooldown $\rightarrow$ "dial-down" (encourage restraint)
- else $\rightarrow$ "neutral"

\textit{post\_contribution\_cooldown}: Pass-through of \texttt{is\_in\_cooldown} boolean

\textit{contribution\_balance}:
- none/light contributions $\rightarrow$ "under" (delegate under-participating)
- active/frequent $\rightarrow$ "over" (delegate dominating)
- else $\rightarrow$ "balanced"

\textit{error\_correction} (null if no issues, otherwise specific guidance):
- If \texttt{off\_topic}: "Prior contribution about [X] diverged from [immediate\_focus]. Re-align to current topic."
- If \texttt{redundancy}: "Repeated [covered content]. Prioritize novel contributions."
- If \texttt{floor\_violation=intrusion}: "Spoke while floor directed to [other]. Respect turn-taking cues."
- If \texttt{hallucination}: "Claimed [unsupported fact]. Ground all claims in EIM/JIT."

\textit{guidance\_summary}: One-sentence synthesis of action outcome and recommendation for next decision.

\textbf{Output Schema}
\texttt{\{last\_action\_outcome, contribution\_pressure, post\_contribution\_cooldown, contribution\_balance, error\_correction, guidance\_summary\}}
\end{prompttext}
\end{subpromptbox}
\end{promptbox}

%% ---------------------------------------------------------------------------
%% A.11 IDEA UNIT SELECTOR (GROUND TRUTH EXTRACTION)
%% ---------------------------------------------------------------------------
\begin{promptbox}[label=prompt:idea-unit-selector]{Idea Unit Selector}
\textbf{Purpose:} Extracts ground-truth idea units from meeting transcripts for a specific participant. Used for evaluation episode construction.

\begin{prompttext}
You are an NLP expert agent specializing in meeting conversation analysis. Goal: extract IDEA UNITS for ONE represented participant from a trimmed transcript table and emit STRICT JSON.

\textbf{Definitions}
- \textbf{Idea Unit}: Exactly one coherent idea: proposal, decision, key fact/metric, stance/opinion, non-trivial question/request
- \textbf{Size}: Usually 1-2 utterances. Split whenever the overall idea changes, even if topic words repeat
- \textbf{Banality Lexicon} (exclude unless inseparable): ok, okay, yeah, yup, thanks, right, got it, sure, mm-hmm, \{vocalsound\}, \{disfmarker\}

\textbf{Cue Types (Preference Order)}
1. \textbf{explicit}: Direct question, address by name, or request to participant
2. \textbf{implicit-handoff}: Topic naturally invites this participant's input (e.g., risk ask leads QA to respond)
3. \textbf{open\_floor}: General invitation (e.g., ``any questions?'')
4. \textbf{self-initiation}: Participant volunteers without explicit prompting

\textbf{Hard Constraints (MUST pass all; otherwise OMIT the Idea Unit)}
1. \textbf{Speaker constraint}: Every \texttt{response\_utterance\_id} MUST be by the represented participant only
2. \textbf{Anchor policy}: Set \texttt{first\_response\_idx} to the MOST substantive response utterance (anchor)
3. \textbf{Cue integrity}: \texttt{cue\_idx} $<$ \texttt{first\_response\_idx} must hold
4. \textbf{Cue proximity} (CRITICAL): When multiple cues qualify, STRONGLY prefer closest cue:
   - Gaps of 1-3 are ideal
   - Gaps of 4-5 are acceptable
   - Gaps of 6+ should be AVOIDED
5. \textbf{Window integrity}: Earliest response must be within next $k$ turns after cue: \texttt{idx} $\in$ (\texttt{cue\_idx}, \texttt{cue\_idx}$+k$]
6. \textbf{Self-initiation fallback}: If natural cue gap $> k$, use immediate previous turn as cue with \texttt{cue\_type}=``self-initiation''

\textbf{Splitting Rule}
If two related but standalone ideas appear (each makes sense alone), split into separate Idea Units.

\textbf{Output Schema}
\texttt{\{
  "idea\_units": [\{
    "unit\_id": "U1",
    "importance\_score": 0.0-1.0,
    "topic\_summary": "<1-2 line idea summary>",
    "response\_utterance\_ids": ["UTT\_ID\_1", ...],
    "first\_response\_idx": <anchor idx>,
    "cue\_utterance\_id": "UTT\_ID\_CUE",
    "cue\_idx": <cue idx>,
    "cue\_type": "explicit|implicit-handoff|open\_floor|self-initiation",
    "anchor\_reason": "<why this cue triggered the idea>",
    "confidence": 0.0-1.0
  \}],
  "coverage\_summary": \{"explicit":0, "implicit-handoff":0, "open\_floor":0, "self-initiation":0\}
\}}
\end{prompttext}
\end{promptbox}

%% file: annotation.tex
\clearpage
\onecolumn
\hypertarget{annotation}{}
\pagestyle{empty}
% --------- Required packages for formatting ---------
\lstset{
  basicstyle=\footnotesize\ttfamily,
  breaklines=true,
  breakatwhitespace=false,
  columns=flexible,
  numbers=none
}

% --------- Define simplified color palette ---------
% Core colors based on Tailwind blue-500
\definecolor{Primary}{RGB}{59, 130, 246}    % Main blue color (Tailwind blue-500)
\definecolor{PrimaryDark}{RGB}{30, 64, 175} % Darker blue for emphasis (Tailwind blue-800)
\definecolor{LightBg}{RGB}{239, 246, 255}   % Very light blue background (Tailwind blue-50)
\definecolor{TextDark}{RGB}{31, 41, 55}     % Dark text color (Tailwind gray-800)
\definecolor{TextMuted}{RGB}{107, 114, 128} % Secondary text color (Tailwind gray-500)

% --------- Header bar ---------
\begin{tikzpicture}[remember picture, overlay]
  \fill[Primary] ([xshift=0cm,yshift=0cm]current page.north west) rectangle ([xshift=\paperwidth,yshift=-0.4cm]current page.north west);
\end{tikzpicture}

\vspace{0.8cm}
\begin{center}
  {\fontsize{22}{26}\selectfont\sffamily\bfseries \textcolor{PrimaryDark}{CiteAssist}}\\[0.2em]
  {\Large\sffamily\scshape \textcolor{TextMuted}{Citation Sheet}}\\[0.8em]
  {\small\sffamily Generated with \href{https://citeassist.uni-goettingen.de/}{\textcolor{Primary}{\texttt{citeassist.uni-goettingen.de}}}
  \CiteAssistCite{}
  }\end{center}

\begin{center}
\vspace{1em}
\begin{tikzpicture}
\draw[Primary, line width=0.6pt] (0,0) -- (\textwidth,0);
\end{tikzpicture}
\vspace{1.2em}
\end{center}

% --------------  BibTeX block  -----------------
\begin{tcolorbox}[enhanced,
                 frame hidden,
                 boxrule=0pt,
                 borderline west={2pt}{0pt}{Primary},
                 colback=LightBg,
                 sharp corners,
                 breakable,
                 fonttitle=\sffamily\bfseries\large,
                 coltitle=Primary,
                 title=BibTeX Entry,
                 attach title to upper={\vspace{0.2em}\par},
                 left=12pt]
\lstset{
    inputencoding = utf8,  % Input encoding
    extendedchars = true,  % Extended ASCII
    literate      =        % Support additional characters
      {á}{{\'a}}1  {é}{{\'e}}1  {í}{{\'i}}1 {ó}{{\'o}}1  {ú}{{\'u}}1
      {Á}{{\'A}}1  {É}{{\'E}}1  {Í}{{\'I}}1 {Ó}{{\'O}}1  {Ú}{{\'U}}1
      {à}{{\`a}}1  {è}{{\`e}}1  {ì}{{\`i}}1 {ò}{{\`o}}1  {ù}{{\`u}}1
      {À}{{\`A}}1  {È}{{\`E}}1  {Ì}{{\`I}}1 {Ò}{{\`O}}1  {Ù}{{\`U}}1
      {ä}{{\"a}}1  {ë}{{\"e}}1  {ï}{{\"i}}1 {ö}{{\"o}}1  {ü}{{\"u}}1
      {Ä}{{\"A}}1  {Ë}{{\"E}}1  {Ï}{{\"I}}1 {Ö}{{\"O}}1  {Ü}{{\"U}}1
      {â}{{\^a}}1  {ê}{{\^e}}1  {î}{{\^i}}1 {ô}{{\^o}}1  {û}{{\^u}}1
      {Â}{{\^A}}1  {Ê}{{\^E}}1  {Î}{{\^I}}1 {Ô}{{\^O}}1  {Û}{{\^U}}1
      {œ}{{\oe}}1  {Œ}{{\OE}}1  {æ}{{\ae}}1 {Æ}{{\AE}}1  {ß}{{\ss}}1
      {ẞ}{{\SS}}1  {ç}{{\c{c}}}1 {Ç}{{\c{C}}}1 {ø}{{\o}}1  {Ø}{{\O}}1
      {å}{{\aa}}1  {Å}{{\AA}}1  {ã}{{\~a}}1  {õ}{{\~o}}1 {Ã}{{\~A}}1
      {Õ}{{\~O}}1  {ñ}{{\~n}}1  {Ñ}{{\~N}}1  {¿}{{?\`}}1  {¡}{{!\`}}1
      {„}{\quotedblbase}1 {“}{\textquotedblleft}1 {–}{$-$}1
      {°}{{\textdegree}}1 {º}{{\textordmasculine}}1 {ª}{{\textordfeminine}}1
      {£}{{\pounds}}1  {©}{{\copyright}}1  {®}{{\textregistered}}1
      {«}{{\guillemotleft}}1  {»}{{\guillemotright}}1  {Ð}{{\DH}}1  {ð}{{\dh}}1
      {Ý}{{\'Y}}1    {ý}{{\'y}}1    {Þ}{{\TH}}1    {þ}{{\th}}1    {Ă}{{\u{A}}}1
      {ă}{{\u{a}}}1  {Ą}{{\k{A}}}1  {ą}{{\k{a}}}1  {Ć}{{\'C}}1    {ć}{{\'c}}1
      {Č}{{\v{C}}}1  {č}{{\v{c}}}1  {Ď}{{\v{D}}}1  {ď}{{\v{d}}}1  {Đ}{{\DJ}}1
      {đ}{{\dj}}1    {Ė}{{\.{E}}}1  {ė}{{\.{e}}}1  {Ę}{{\k{E}}}1  {ę}{{\k{e}}}1
      {Ě}{{\v{E}}}1  {ě}{{\v{e}}}1  {Ğ}{{\u{G}}}1  {ğ}{{\u{g}}}1  {Ĩ}{{\~I}}1
      {ĩ}{{\~\i}}1   {Į}{{\k{I}}}1  {į}{{\k{i}}}1  {İ}{{\.{I}}}1  {ı}{{\i}}1
      {Ĺ}{{\'L}}1    {ĺ}{{\'l}}1    {Ľ}{{\v{L}}}1  {ľ}{{\v{l}}}1  {Ł}{{\L{}}}1
      {ł}{{\l{}}}1   {Ń}{{\'N}}1    {ń}{{\'n}}1    {Ň}{{\v{N}}}1  {ň}{{\v{n}}}1
      {Ő}{{\H{O}}}1  {ő}{{\H{o}}}1  {Ŕ}{{\'{R}}}1  {ŕ}{{\'{r}}}1  {Ř}{{\v{R}}}1
      {ř}{{\v{r}}}1  {Ś}{{\'S}}1    {ś}{{\'s}}1    {Ş}{{\c{S}}}1  {ş}{{\c{s}}}1
      {Š}{{\v{S}}}1  {š}{{\v{s}}}1  {Ť}{{\v{T}}}1  {ť}{{\v{t}}}1  {Ũ}{{\~U}}1
      {ũ}{{\~u}}1    {Ū}{{\={U}}}1  {ū}{{\={u}}}1  {Ů}{{\r{U}}}1  {ů}{{\r{u}}}1
      {Ű}{{\H{U}}}1  {ű}{{\H{u}}}1  {Ų}{{\k{U}}}1  {ų}{{\k{u}}}1  {Ź}{{\'Z}}1
      {ź}{{\'z}}1    {Ż}{{\.Z}}1    {ż}{{\.z}}1    {Ž}{{\v{Z}}}1  {ž}{{\v{z}}}1
      % ¿ and ¡ are not correctly displayed if inconsolata font is used
      % together with the lstlisting environment. Consider typing code in
      % external files and using \lstinputlisting to display them instead.      
  }
\begin{lstlisting}
@inproceedings{khan-etal-2026-meetingdelegation,
  address={Budapest, Hungary},
  author={Khan, Muneeb and Kirstein, Frederic and Ruas, Terry and Gipp, Bela},
  booktitle={The 2026 Conference on Empirical Methods in Natural Language Processing},
  month={oct},
  publisher={Association for Computational Linguistics},
  title={Speak for Me: Giving LLMs the Situational Awareness to Participate in a Meeting},
  year={2026}
}
\end{lstlisting}
\end{tcolorbox}

% % ------ Footer with subtle design element ------
% \vfill
% \begin{tikzpicture}
% \draw[Primary!40, line width=0.4pt] (0,0) -- (\textwidth,0);
% \end{tikzpicture}
% \begin{center}
% \small\sffamily\textcolor{TextMuted}{Generated \today}
% \end{center}